\documentclass[letterpaper,conference]{ieeeconf}

\IEEEoverridecommandlockouts                              % This command is only needed if 
\usepackage{graphicx}
\usepackage{amssymb}  % assumes amsmath package installed
\usepackage{hyperref,tcolorbox}
\usepackage{subcaption}
\usepackage{enumerate}
\usepackage{cite}
\usepackage{flushend}
\usepackage{amsmath,bm}
\usepackage[algo2e]{algorithm2e} 
\usepackage{soul} % strikethrough

\newcommand\norm[1]{\left\lVert#1\right\rVert}

\usepackage{scrextend}

\newif\ifNotAnonymous
\NotAnonymoustrue
\newif\ifresubmissionblue
\resubmissionbluefalse

\newif\ifmodification
\modificationfalse

\newif\iffull
\fulltrue
\newif\ifshort
\shortfalse
\newif\ifNotEricTrim
\NotEricTrimfalse
\newif\ifEricTrim
\EricTrimtrue
\title{\LARGE \bf
Configuration-Dependent Robot Kinematics Model and Calibration
}

\def\PH{{{\bf P},{\bf H}}}
\def\P{{\bf P}}
\def\H{{\bf H}}
\def\supk{^{(k)}}
\def\sq{^2}
\def\tr{^\top}

\def\ERP{\Theta_{\P}}
\def\ERH{\Theta_{\H}}
\def\ERTH{\Theta_{\PH}}

\ifNotAnonymous{
\author{Chen-Lung Lu$^\dagger$\thanks{$^\dagger$ Rensselaer Polytechnic Institute, Troy, NY. Emails: {\tt luc6@rpi.edu heh6@rpi.edu agung@ecse.rpi.edu wenj@rpi.edu}}, Honglu He$^\dagger$, Agung Julius$^\dagger$, John T.~Wen$^\dagger$
}
}\fi
\usepackage{titlesec}
\titlespacing{\section}{0pt}{*}{*} % Adjusts section spacing
\titlespacing{\subsection}{0pt}{*}{*} % Adjusts subsection spacing
\titlespacing{\subsubsection}{0pt}{-2pt}{-2pt} % Adjusts subsubsection spacing
\titlespacing{\paragraph}{0pt}{*}{*} 

\begin{document}
\maketitle
\thispagestyle{empty}
\pagestyle{empty}

%%%%%%%%%%%%%%%%%%%%%%%%%%%%%%%%%%%%%%%%%%%%%%%%%%%%%%%%%%%%%%%%%%%%%%%%%%%%%%%%
\begin{abstract}
\ifmodification{\color{red} \st{
Accurate robot kinematics is critical for precise tool placement using joint control of an articulated robot. Traditional kinematics calibration using Denavit-Hartenberg parameters is sensitive to joint-axis directions. Product-of-Exponentials (POE) parameterization is more robust but is non-minimal. 
Robot kinematics is also affected by the pose of the robot in different parts of the workspace due to joint tolerance, especially for massive payloads.
%
%A fixed set of rigid robot kinematics cannot maintain the same level of tool accuracy over the entire workspace due to joint tolerance, especially under massive payloads. 
This paper compares multiple POE-based kinematics calibration methods.
%for robots in a robotic welding cell. 
We propose a configuration-dependent parameterization which is shown to have the best overall performance. In addition, for a traditional 6-DoF elbow manipulator, the configuration dependency is mostly on the shoulder and elbow angles. We show that the configuration dependence of the kinematics parameters may be adequately captured by these two angles. 
%We demonstrate the utility of this method in improving the robot tool path trajectory tracking accuracy for two Motoman robots with motion capture of the tool frame as the calibration baseline.
}}\fi
\ifresubmissionblue{\color{blue}\fi 
Accurate robot kinematics is essential for precise tool placement in articulated robots, but non-geometric factors can introduce configuration-dependent model discrepancies. This paper presents a configuration-dependent kinematic calibration framework for improving accuracy across the entire workspace. Local Product-of-Exponential (POE) models, selected for their parameterization continuity, are identified at multiple configurations and interpolated into a global model. Inspired by joint gravity load expressions, we employ Fourier basis function interpolation parameterized by the shoulder and elbow joint angles, achieving accuracy comparable to neural network and autoencoder methods but with substantially higher training efficiency. Validation on two 6-DoF industrial robots shows that the proposed approach reduces the maximum positioning error by over 50\%, meeting the sub-millimeter accuracy required for cold spray manufacturing. Robots with larger configuration-dependent discrepancies benefit even more. A dual-robot collaborative task demonstrates the framework’s practical applicability and repeatability.
\ifresubmissionblue}\fi

\end{abstract}

%%%%%%%%%%%%%%%%%%%%%%%%%%%%%%%%%%%%%%%%%%%%%%%%%%%%%%%%%%%%%%%%%%%%%%%%%%%%%%%%
\section{INTRODUCTION}
\label{sec:intro}

% \jw{Use more general widely-cited references, preferably not our own, unless absolutely necessary.}
Industrial robot arms are increasingly utilized in manufacturing processes demanding high accuracy, such as cold spray \cite{chen2008automated}, and deep rolling \cite{chen2020robotics}.
% , and surface grinding \cite{zhu2020robotic}.
To ensure that the robot tool center point (TCP) follows the desired path using joint control, accurate robot kinematic parameters are needed.
%Robots offer certain advantages over the traditional approach with computer numerical control (CNC) machines, including larger workspaces and lower unit costs. However, robot arms usually have lower accuracy in terms of controlling the tool center point (TCP) than CNC, especially with heavier payload. 
%which makes the spread of robot arms in high-accuracy-demanding processes difficult. 
%Performing 
%Robot calibration is a key step 
%to obtain accurate kinematic models that close the gap between the ideal kinematic model and the real robots is necessary.
%
Using kinematics parameters directly from the vendor data sheet often results in TCP errors beyond the vendor specification.  The kinematics error may be caused by geometrical and non-geometrical factors \cite{kana2022fast}. Geometrical factors, such as misalignment in joint axes during robot production 
\ifNotEricTrim{and mismatch in encoder position and joint angle readings}\fi, 
are typically not dependent on the arm configuration (e.g., out-stretched vs.~folded).
%universal in all robot configurations, and may be captured in a  single calibrated model. 
The non-geometrical factors, such as link and joint flexibility, motor backlash, etc., depend on the robot configuration due to the different loading conditions on the robot joints.  

\begin{figure}[ht]
    \centering
    \includegraphics[width=0.5\textwidth]{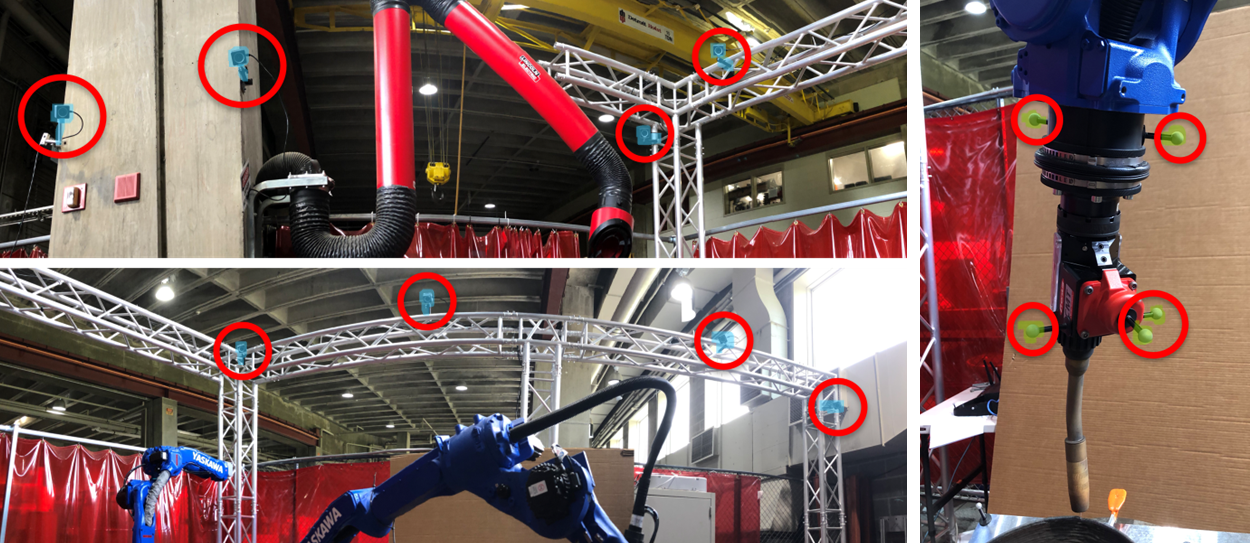}
    \caption{Experimental robot calibration setup using the OptiTrack motion capture system.}
    \label{fig:optitrack_setup}
\end{figure}

Robot calibration is the process of using external TCP measurements (considered as the ground truth) and robot joint angle readings to fit a kinematic model, so that joint angles can predict the TCP location sufficiently accurately during runtime.  Fig.~\ref{fig:optitrack_setup} shows an example of the external TCP measurement using a motion capture system from our robot welding testbed. 

%\jw{This paragraph should go before the previous paragraph.}
There are four key steps in robot calibration: modeling, measurement, identification, and compensation \cite{roth1987overview}. The critical aspect lies in effectively parameterizing the robot kinematic model. 
Common approaches include the Denavit-Hartenberg (DH) parameterization 
% \cite{Hayati1985ImprovingTA,Hayati1983robot,Veitschegger1986robot,Zhuang1992cpc} 
\cite{Hayati1983robot} 
and the Product-of-Exponentials (POE) model 
% \cite{okamura1996kinematic,he2010kinematic,he2014kinematic,li2014identifiability,Wang2014calibration,li2016poe}.
\cite{okamura1996kinematic}.
%\jw{Do we need these many references for DH and POE?}
Although the DH model is widely adopted in both industrial and academic settings, it suffers from non-smooth parameterization and the presence of singularities, particularly in configurations with parallel joint axes \cite{zhuang1993error}. To mitigate these issues, \cite{Hayati1985ImprovingTA,Veitschegger1986robot} introduced an additional parameter for parallel or near-parallel joint axes, a strategy adopted by most subsequent works using the DH parameters, such as \cite{Hsiao2020positioning,jang2001calibration,chen2022kinematic}. However, not only this approach is non-minimal, it is ad-hoc and further complicates the calibration process \cite{Park1994kinematic,HE2024local}.
The POE model offers a smooth and continuous parameterization, but is non-minimal. 
%representing the robot arm as a series of link vectors and joint rotation axes. The POE model primarily employs twist coordinates to describe joint relationships \cite{okamura1996kinematic, he2010kinematic, he2014kinematic}. Nevertheless, the additive error model of the standard POE parameterization includes 6 parameters for each revolute joints which is clearly non-minimal. 
Recent works \cite{li2016poe} imposes additional constraints to remove the redundancy in the POE parameterization for robot calibration.

%Recent works have identified redundancy \cite{li2016poe} and states that all the possible joint configuration is the tangent bundles of a sphere. \cite{yang2014minimal} assumes small error in the twist coordinate and thus constraints the link origin on the initial tangent plane. But the additive error model needs further normalization for joint axis. \cite{li2016poe,li2014identifiability} consider a multiplicative adjoint error on the nominal twist coordinate. The possible adjoint error twist in this work exclude the isotropy subgroup to achieve minimality. However, using twist coordinates and adjoint error has less physical interpretations considering the configuration dependent kinematic model. \jw{why?}

Calibration for the geometrical factors has long been studied \cite{Hayati1985ImprovingTA,okamura1996kinematic} and is now routinely applied.  
% maybe comment out
Compensating for the non-geometric factors to achieve high accuracy over the entire workspace has received comparatively less attention.  
%
%\jw{Is this statement true? Is there a reference?}
%is still a worth studying topic. 
%\jw{Describe what calibration means: find the relationship between robot joint angle measurements and the tool position and orientation.  The ground truth TCP is measured using some external means (include citations of laser marker, motion capture, as shown in Fig.~\ref{fig:optitrack_setup} from our robot welding testbed, etc.)
%}
Many calibration methods for non-geometric factors focus on identifying joint flexibility
%Some of the work studied the effect on joint flexibility which dominants all other non-geometric factors, such as
\cite{Zhou2014Simultaneous,Kamali2016elasto,wang2020improvement}. 
%While the joint torques compensating for the robot link and tool loads, the torques further apply on the joint spring and cause deflections. 
These papers model joints as linear torsional springs and estimate the compliance coefficients of each joint. The Jacobian matrix of the tool offset with respect to the compliance coefficients is derived for optimization. 
Zhou et~al.~\cite{Zhou2014Simultaneous} %developed an algorithms to 
simultaneously calibrates the geometric factors and the joint compliance coefficients. 
%
% \jw{What is the algorithm that Kamali used?}
Kamali et~al.~\cite{Kamali2016elasto} developed a %creative 
cable system to apply various load to the robot to increase data collection efficiency. The parameters are optimized using damped least-square at different poses and loads. 
Although estimating the joint compliance is a promising approach, it requires %center of mass, link mass, tool load information, or having 
the ability to measure the joint torques or the wrench applied on the tool, which may not be readily available.
%These information or measuring sensors are sometimes hard to retrieved or not readily available, which limit the usage of these approaches.
Several studies focus on model-free, configuration-dependent calibration methods, where the nominal kinematic parameters are first calibrated, followed by a data-driven model relating the residual position and orientation errors as a function of the joint configuration and tool payload \cite{Hsiao2020positioning,landgraf2021hybrid,zhu2019positioning,zhao2019system}. The configuration to residual error function is mostly approximated by a multi-layer feedforward Neural Network (NN) or radial basis functions (RBF). 
%Besides back-propagation, NN is also trained in different way such as Butterfly and Flower Pollination Algorithm \cite{cao2022robotcal}, teaching-learning based optimization \cite{le2020robot}, or using Differential Evolution in Meta-learning style \cite{jiang2020absolute}, in order to avoid local minimum. 
The NN prediction can further be implemented on top of the joint stiffness model to compensate for other non-geometric factors \cite{wang2020improvement,nguyen2019calibration,le2020robot}. 
With extensive training, these black-box models can increase the tool location accuracy, 

%on top of the geometrical calibration, we cannot get any insight on how the joint configuration affects the kinematics parameters. Jiang et. al. \cite{jang2001calibration} propose a method to predict the kinematics parameters variance instead of position residual error using Radial basis function (RBF). The work shows the parameter variance across different configuration. However, neither using NN nor RBF leverages the insight on the dynamic model of the robot, which is readily available. \jw{Not sure what the preceding sentence means.} 
%

In this work, we adopt the POE model for robot parameterization. By requiring the identified model to be close to the nominal POE parameters, we remove the redundancy in the parameterization. Using local gradient search with data collected from clusters of robot configurations, we identify a set of POE parameters {\em near} the nominal parameters at each cluster. The full kinematics is constructed through interpolation of these local kinematic models.  We found that the most effective interpolation scheme is to use a set of Fourier basis functions parameterized on the robot shoulder and elbow angles.  This basis is motivated by the expression of the joint gravity load \cite{murray1994mathematical}.  For comparison, we also train a NN directly from the collected data. The performance is comparable, though the identification of the Fourier-basis model is much more efficient as no extensive training is required.  We demonstrate our method on two separate 6R Motoman robots in our welding testbed.  
\ifNotEricTrim{An OptiTrack motion tracking system is used for the tool frame measurements.}\fi  Two sets of data are collected, one for model identification and one for performance evaluation.  The test results show that the Fourier-basis method is able to reduce the maximum error by more than 50\% to within 1~mm. \ifresubmissionblue{\color{blue}\fi To summarize, the main contribution of this paper are: \vspace{-0.5mm}
\begin{itemize}
    \item A configuration-dependent kinematic calibration framework based on local minimal POE model identification and interpolation, which improves accuracy across the full workspace.
    \item A Fourier basis function approximation strategy parameterized by shoulder and elbow joint angles, offering an efficient and physically motivated alternative to pure data-driven methods such as neural networks.
    \item Experimental validation on two 6R industrial robots, demonstrating significant improvements in accuracy in reducing the maximum position error by more than 50\% compared to the nominal model. The calibrated models were further evaluated in a dual-robot collaborative contact test, confirming their practical applicability.
\end{itemize}
\ifresubmissionblue}\fi

The paper is organized as follows. Section \ref{sec:model_calibration} describes the robot kinematics and kinematic model calibration. Section \ref{sec:CDC} introduces our configuration-dependent kinematic model and the corresponding calibration frameworks. In section \ref{sec:exp}, we show the experimental results and demonstrate the enhanced accuracy using the proposed methodology. 
%Additionally, we demonstrate that the calibrated robot could be utilized in a trajectory tracking application where accuracy is a top requirement. Section \ref{sec:conclusion} presents a summary and future research directions.

%%%%%%%%%%%%%%%%%%%%%%%%%%%%%%%%%%%%%%%%%%%%%%%%%%%%%%%%%%%%%%%%%%%%%%
\section{Robot Kinematics Model and Calibration}
\label{sec:model_calibration} 

\subsection{POE Forward Kinematics Model}

\def\rot{{\sf R}}
\def\calE{{\mathcal E}}
\def\calO{{\mathcal O}}
\def\calR{{\mathcal R}}

\begin{figure}[ht]
    \centering    \includegraphics[width=0.25\textwidth]{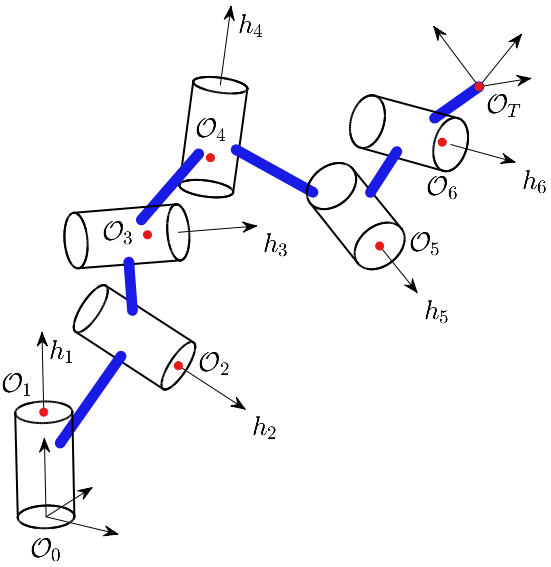}
    \caption{Kinematic diagram of a 6R robot. %\jw{show a general 6-dof arm with no intersections, parallel axes, etc.}
    }
    \label{fig:robot_diag}
\end{figure}

Consider an $n$-joint articulated robot arm with all revolute joints as shown in Fig.~\ref{fig:robot_diag}.  Denote the reference (world) coordinate frame by an orthonormal frame and the origin: $(\calE_0,\calO_0)$.   
The $i$th frame, $\calE_i$, is the rotation of the $i-1$th frame, $\calE_{i-1}$, about the joint axis, $\vec h_i$, over the joint angle $q_i$.  The joint axis in the $i-1$th frame (and the $i$th frame) is a constant vector, denoted by $h_i$. 
%
%Denote the $i$th joint axis by the unit vector $h_i$. 
Denote $\calE_i$ represented in $\calE_j$ by the rotation matrix $R_{ji}$.  
The relative rotation between two consecutive frames is the rotation about the joint axis:
\begin{equation}
    R_{i-1,i}= {\sf R}(h_i,q_i) 
\end{equation}
where 
%${\sf R}$ is the rotation about $h_i$ over the joint angle $q_i$.  It may be computed using the Euler-Rodrigues formula or the matrix exponential 
${\sf R}(h_i,q_i)= e^{h_i^\times q_i}$, $(\cdot)^\times$ is the skew symmetric matrix representing the cross product. The tool coordinate frame $\calE_T$ represented in $\calE_n$ is denoted by $R_{nT}$.

The origin of the $i$th frame, $\calO_i$, may be arbitrarily chosen along $h_i$.  Denote the link vector from $\calO_{i-1}$ to $\calO_i$ by $p_{i-1,i}$ which is a constant vector in $\calE_{i-1}$. The same notation applies for the link vector from $\calO_{n}$ to $\calO_T$, denote by $p_{n,T}$.
% \begin{equation}
%     p_{i-1,i} = \calO_i-\calO_{i-1}
% \end{equation}
Define a homogeneous transformation from the $i-1$th frame to the $i$th frame as 
\begin{equation}
    T_{i-1,i} = \begin{bmatrix}
        R_{i-1,i} & p_{i-1,i} \\ 0 & 1 
    \end{bmatrix}.
    \label{eq:fwdkin}
\end{equation}
Denote the robot tool frame (end effector position and orientation) by $(p_{0T},R_{0T})$.  Then the robot forward kinematics (mapping joint angles $\{q_i\}$ to the tool frame) is given by
\begin{equation}
    T_{0T}(q) = T_{01}(q_1) T_{12}(q_2) \ldots T_{n-1,n}(q_n) T_{nT}
\label{eq:fwd}
\end{equation}
where $q$ is a $n\times 1$ vector consisting of all joint angles. Note that $T_{nT}$ is a constant transform: 
\begin{equation}
    T_{nT} = \begin{bmatrix}
        I & p_{nT} \\ 0  & 1 
    \end{bmatrix}.
\end{equation}
The robot forward kinematics is completely parameterized by the joint axes $\{h_i\}$ and link axes $\{p_{i-1,i}\}$.  We can put them in the matrix form:
\begin{equation}
    {\bf H} = \begin{bmatrix} h_1 &\ldots & h_n
    \end{bmatrix}, \quad 
    {\bf P} = \begin{bmatrix} p_{01} &\ldots & p_{nT} \end{bmatrix}.
\end{equation}
The tool frame in forward kinematics \eqref{eq:fwdkin} may be rewritten as below to highlight the dependence on the kinematic parameters:
\begin{equation}
    \begin{split}
        T_{0T}(q;{\bf P},{\bf H}) &= T_{01}(q_1;p_{01},h_1)\ldots \\
    &T_{n-1,n}(q_n;p_{n-1,n},h_n) T_{n,T}(p_{nT}).
    \end{split}
    \label{eq:T_0T}
\end{equation}
The kinematic calibration problem is to use multiple measurements of the joint angles $q$ and the corresponding tool frames $T_{0T}$ to estimate $(\bf P,\bf H)$.  We assume the robot manufacturer has provided a set of nominal parameters: 
\begin{equation}
    \bar {\bf H} = \begin{bmatrix} \bar h_1 &\ldots & \bar h_n
    \end{bmatrix}, \quad 
    \bar {\bf P} = \begin{bmatrix} \bar p_{01} &\ldots & \bar p_{nT} \end{bmatrix}.
    \label{eq:HP}
\end{equation}
We will first estimate $(\bf P,\bf H)$ using data near a specified configuration (i.e., a given joint angle $q$).   The process is then generalized to multiple configurations throughout the workspace.  For an arbitrary configuration, we can interpolate from estimated $(\bf P,\bf H)$ values in nearby configurations.  We shall see that the interpolation algorithm affects the overall accuracy of the calibration result.

\subsection{A Minimal POE Model}
\label{sec:minimalPOE}

\begin{figure}[ht]
    \centering
    \includegraphics[width=0.27\textwidth]{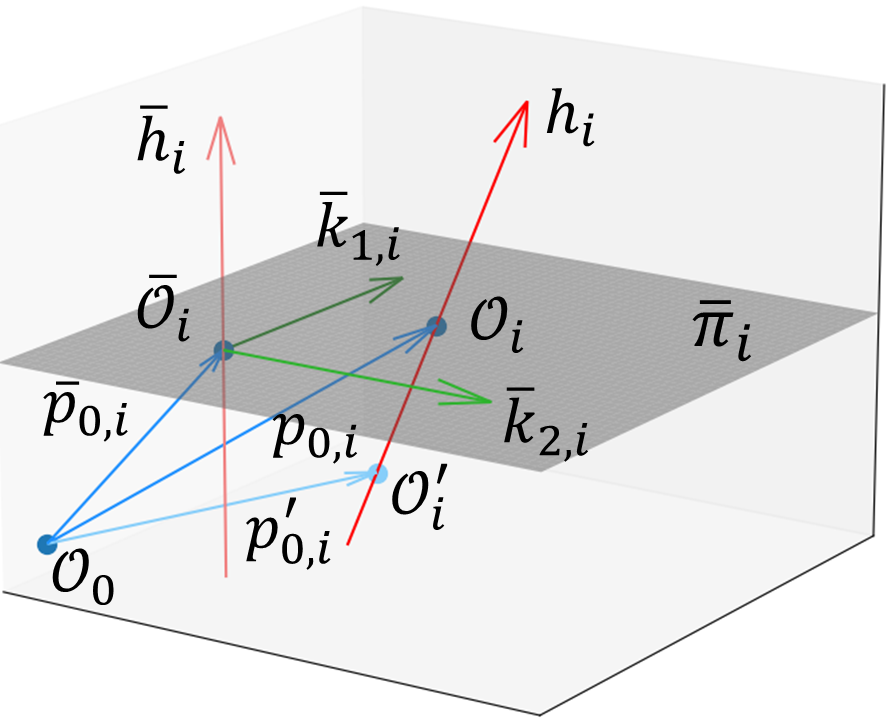}
    \caption{Minimal POE parameterization based on a set of nominal POE parameters. 
    \ifresubmissionblue{\color{blue}\fi 
    Redundancy in the location of $\calO_i$ is removed by constraining it to lie in the plane perpendicular to the nominal joint axis through the nominal joint origin.
      % By constraining the identified origin $\calO_i$ to be the intersection of the identified axis $(h_i)_o$ with $\bar \pi_i$, the redundancy can be removed. We then described the calibrated model as an offset from the nominal parameters.
      \ifresubmissionblue}\fi}
    \label{fig:axe_geo}
\end{figure}

The POE parameterization is not minimal: The $i$th frame origin $\calO_i$ may be placed anywhere on the $i$th joint axis, and the $i$th joint axis $h_i$ is a unit vector, $\norm{h_i}=1$. To identify a unique POE model, we constraint the identified parameters to be {\em close} to the nominal parameters (Fig.~\ref{fig:axe_geo}).  
Denote $\bar h_i$ in the base frame as $(\bar h_i)_o:=R_{0,i-1}\bar h_i$. Define $\bar\pi_i$ be the plane perpendicular to $(\bar h_i)_o$ and intersecting $\bar\calO_i$, the $i$th origin in the nominal model.  We require the identified origin $\calO_i$ to be the intersection of the identified axis $(h_i)_o$ with $\bar \pi_i$.   
%The spatial line along $(\bar h_i)_o$ through $\bar\calO_i$ is given by $\bar p_{0i}+\bar\alpha (\bar h_i)_0$ for $\bar\alpha\in\mathbb{R}$.  
Let $p'_{0i}$ be any vector from the robot base to the line $(h_i)_o$.  Then $\calO_i$ is given by the tip of the vector
\begin{equation}
    p_{0i} = p'_{0i}+ \frac{(\bar p_{0i}-p'_{0i})^T(\bar h_i)_o}{(h_i)_o^T (\bar h_i)_o} (h_i)_o.
\end{equation}
Given the joint angle $q$, we can iteratively recover the constant vectors in $\bf H$ and $\bf P$ in \eqref{eq:HP} from $(h_i)_o$ and $p_{0i}$: 
\begin{subequations}
\setlength{\jot}{2pt}  % Redu
\begin{align}
    h_i &= R_{0,i-1}^T (h_i)_o \\
    R_{0i}&= R_{0,i-1}e^{h_i^\times q_i} \\
    p_{i-1,i} &= R_{0,i-1}^T (p_{0i} - p_{0,i-1})
\end{align}    
\label{eq:HP_id}
\end{subequations}
for $i=1,\ldots,n+1$, with $\calO_{n+1}=\calO_T$.

%First, write $h_i$ as two consecutive rotations of $\bar h_i$ with rotation axes perpendicular to $\bar h_i$.  
%

% \jw{Include a figure here to illustrate.}

% Let $(k_{1,i},k_{2,i})$ be an orthonormal basis for $\bar\pi_i$ (they are easily determined from $\bar h_i$). Then we can parameterize $h_i$ by two rotation angles: 
% \begin{equation}
%     h_i = \rot(k_{1,i},\theta_i)\rot(k_{2,i},\phi_i)\bar h_i.
%     \label{eq:joint_axis_error}
% \end{equation}
% Write the vector from $\bar\calO_{i}$ to $\calO_i$, represented in the $i$th frame, in terms of the same basis functions of $\bar\pi_i$, $v_i k_{1,i} + w_i k_{2,i}$ where $(v_i,w_i)$ are constants to be determined. Then the link vectors may be parameterized by two distances:
% \begin{equation}
% \begin{aligned}  
% \!\!\! p_{i-1,i} &=\bar p_{i-1,i} + (v_i k_{1,i} + w_i k_{2,i})
% - (v_{i-1} k_{1,i-1} + w_i k_{2,i-1}) \\
%         & \quad i=1,\ldots,n+1,
%  \end{aligned}
%     \label{eq:link_O}
% \end{equation}
% where the $\calO_{n+1}=\calO_T$. Note that since the base and tool origins are fixed, $v_0=w_0=0$ and $v_{n+1}=w_{n+1}=0$.

%%% 
% Add details for non-zero $q$
%%%

\subsection{Circular-Point Analysis (CPA)}
\label{sec:cpa}

First consider a simple and intuitive POE parameter identification method circular-point analysis (CPA) \cite{mooring1991calibration}.
% , which we call the circular-point analysis (CPA).
% to identify $(\bf H,\bf P)$ at a specific configuration.   
Move the robot to fixed location corresponding to the joint angle $q$.  Then command the $i$th joint to $K$ different values and record the corresponding TCP positions:
\begin{equation}
    p_{0T}^{(k)} = p_{0i} + e^{(h_i)_o^\times q_i^{(k)}} (p_{iT})_o,
\end{equation}
with $k=1,\ldots, K$.  Fit these positions, $\{p_{0T}^{(k)}\}$, to a circle using a standard 3D circle fitting algorithm \cite{fitcircle2000,meshlogic} to find $(h_i)_o$ and $p'_{oi}$ (normal and center of the circle).   The identified POE parameters may then be extracted using \eqref{eq:HP_id}. 
The CPA method is conceptually simple but each joint must move to multiple locations for a given $q$.  For multiple configurations, the process could be time consuming.

\subsection{Nonlinear Least Square (NLS) Estimate}
\label{sec:NLS}

\def\Tgt{T_{0T}^{*\supk}}

An alternate approach is to estimate $(\PH)$ by solving a nonlinear least square (NLS) problem.  
From Section~\ref{sec:minimalPOE}, we can parameterize $p_{i-1,i}$ and $h_i$ by two parameters each:
\begin{subequations}
\setlength{\jot}{2pt} 
\begin{align}
     h_i &= \rot(k_{1,i},\theta_i)\rot(k_{2,i},\phi_i)\bar h_i      \label{eq:joint_axis_error}
     \\
    \!\!\! p_{i-1,i} &=\bar p_{i-1,i} \!+\! (v_i k_{1,i} \!+ w_i k_{2,i})\! -\! (v_{i-1} k_{1,i-1}\! + w_{i-1} k_{2,i-1}) 
    \label{eq:link_O}
    \end{align}
\end{subequations}
where $(k_{1,i},k_{2,i})$ are two orthogonal vectors in $\bar\pi_i$, and $(\theta_i,\phi_i)$ parametrize $h_i$ relative to $\bar h_i$ and $(v_i,w_i)$ parameterizes the location of $\calO_i$ relative to $\bar O_i$.  Note that since the base and tool origins are fixed, $v_0=w_0=0$ and $v_{n+1}=w_{n+1}=0$.  
%There are therefore $2n$ kinematic parameters.  
Define
\begin{equation}
\setlength{\jot}{2pt}
\begin{aligned}    
    \ERP &= [v_1,w_1,\ldots,v_n,w_n]\tr\\ 
    \ERH &= [\theta_1,\phi_1\ldots,\theta_n,\phi_n]\tr \\
    \ERTH &= \begin{bmatrix}
        \ERP \\ \ERH
    \end{bmatrix}.
\end{aligned}
\label{eq:paramvec}
\end{equation}
There are  $4n$ total parameters.
The metric for the NLS is 
\begin{equation}
\label{eq:nls_obj}
    \min_{\ERTH} \frac12 \sum_{k=1}^K \norm{T_{0T}\left(q^{(k)},\P(\ERP),\H(\ERH)\right)- T_{0T}^{*\supk}}^2.
\end{equation}
where $K$ is the total collected data samples, $\Tgt$ is the ground truth measurement at sample $k$. 
For the error norm, we choose a weighted combination of the position error and the orientation error:
\begin{equation}
\norm{T_1-T_2}\sq = \norm{p_1-p_2}\sq + a \norm{R_1R_2\tr-I_3}_F^2
\label{eq:trans_norm}
\end{equation}
where $a$ is a weighting constant and the orientation error norm is equivalent to
% {\color{red} This is incorrect, it should be $a \sin^2\left(\frac{\theta}2\right)$}
$8\mbox{sin}^2(\frac{\theta}{2})$, $\theta$ is the equivalent angle of  $R_1R_2\tr$.   
% {\color{red} For 1deg error in $\theta$, we have $8\sin(\theta/2)^2=6.1\cdot 10^{-4}$.  So $a$ should be $1641$.  Is this what you used?}
%In our manufacturing application, cold spraying, requires sub-millimeter and sub-degree accuracy. To balance the impact of 1 mm and 1-degree errors in our loss function, we set the weighting factor to $a=1/(8\sin^2(\frac{\pi}{360}))=1641$.
We apply the gradient descent algorithm, starting with the nominal POE values, i.e., $\Theta_P=\Theta_H=0$. The detail of the gradient computation is presented in \iffull{~\ref{app:gradient}.}\fi \ifshort{\cite{paper_full}. }\fi For a given configuration $q$, we collect a set of joint angles and task frame data near $q$ and then estimate the POE parameters for that local data cluster.

\section{Configuration Dependent Calibration (CDC)}
\label{sec:CDC}

The NLS method finds a set of POE parameters, $\ERTH$, around a given configuration using nearby data points.  We may repeat the process throughout the robot workspace to obtain multiple sets of POE parameters, $\ERTH^{(\ell)}$, $\ell=1,\ldots,L$, corresponding to the centroid of the cluster $q^{(i)}$.  To find the POE parameters at an arbitrary configuration $q$, we may apply an interpolation method (e.g., nearest neighbor, linear interpolation, cubic interpolation, etc.) based on the identified cluster parameters.  

For elbow type of manipulators, $\ERTH^{(\ell)}$ shows a strong dependence of the shoulder and elbow angles, $(q_2,q_3)$ due to the moment arm effect. 
%We further calculate the tool frame error using the nominal (vendor-provided) kinematics data.  The positional error and the joint angles correlation is shown in Table\ref{tab:error_joint_cc}. Not surprisingly, the most pronounced dependence is on the shoulder and elbow angles, $(q_2,q_3)$, as the moment-arm effect is most pronounced with a longer link and mass. Therefore we collect training clustered at different configuration, i.e.  $(q_2,q_3)$ sets, and obtained the $\ERTH$ and thus ($\PH$) parameters using NLS. 
Note that if we have longer extension from the robot wrist, there will be increasing dependence on the wrist angles.  In this case, we may write the POE parameters as 
\begin{equation}
    \ERTH(q) = f_{\ERTH} (q_2,q_3).
    \label{eq:fq2q3}
\end{equation}
We will first consider a basis function expansion approach to approximate $f_{\ERTH}$ as a linear combination of a chosen basis set, and use the estimated configuration-dependent kinematic parameters $(q^{(\ell)},\ERTH^{(\ell)})$ to learn the expansion coefficients.  We will also consider a direct blackbox approximation using a multilayer neural network, and train using the identified POE parameters.  

%between the error and the spherical joint angels. We need to consider the spherical joints configuration in this case.

%Given that joints 2 and 3 have the most significant impact on the robot forward kinematics, 
% Once the error parameters $\ERTH$ for the sampled clustered are obtained, we can compute the $\ERTH$ for any $(q_2,q_3)$ set by interpolating the estimated $\ERTH$ at nearby sampled $(q_2,q_3)$ and finally acquired $(\PH)$ using Eq. \ref{eq:link_error},\ref{eq:joint_axis_error}.
% We can also learn a function to predict the error parameters $\ERTH$ from the configuration. 
%Assume that the error parameters $\ERTH$ is a function of $(q_2,q_3)$, the configuration dependent parameters is as followed.
% \begin{equation}
%     \ERTH = f_{\ERTH} (q_2,q_3)
% \end{equation}
% We proposed three way to learn the function $f_{\ERTH} (q_2,q_3)$.

\subsection{Basis Function Approximation}
\label{sec:cdc_fourier}
%As stated in the previous section, inspired by the symbolic expression of the gravity load matrix, we would like to approximate the configuration dependent function with FBF. 

Write the parameter vector $f_{\ERTH}$  \eqref{eq:fq2q3} as a linear combination of $m$ basis functions, collected in  a column vector $\beta(q_2,q_3)\in\mathbb{R}^m$:
\begin{equation}
    f_{\ERTH}(q_2,q_3)=A \beta(q_2,q_3)
    \label{eq:basisexpansion}
\end{equation}
where $A$ is a constant coefficient matrix. Using the identified POE parameters from $L$ clusters, we have 
\begin{equation}
\setlength{\jot}{2pt}
    \begin{split}
        \underbrace{\begin{bmatrix}
        \Theta_\PH(q^{(1)}) & \ldots & \Theta_\PH(q^{(L)})
        \end{bmatrix}}_{{\bf C}}
        =  \\
        A
        \underbrace{\begin{bmatrix}
            \beta(q_2^{(1)},q_3^{(1)}) & \ldots & \beta(q_2^{(L)},q_3^{(L)})
        \end{bmatrix}}_{\bf B}.    
    \end{split}
\end{equation}
The least square estimate of $A$ is:
\begin{equation}
    A = {\bf C} {\bf B}^+
    \label{eq:A}
\end{equation}
where ${\bf B}^+$ is the Moore-Penrose pseudo-inverse of ${\bf B}$. Then given any $q$, we can compute $\ERTH$ using \eqref{eq:basisexpansion} and \eqref{eq:A}.
One may use a generic basis set such as the radial basis functions (RBF) \cite{Buhmann2000radial}.  For an elbow-type manipulator, the configuration dependence is likely due to the gravity load on the elbow and shoulder joints.  In this case, a reasonable basis is the Fourier basis based on the expansion of the gravity load \ifresubmissionblue{\color{blue}\fi \cite{murray1994mathematical} \ifresubmissionblue}\fi:
\def\sin{\mbox{sin}}
\def\cos{\mbox{cos}}
\def\s{{s}}
\def\c{{c}}
\def\twoqtwo{{\scriptstyle 2q_2}}
\def\twoqthree{{\scriptstyle 2q_3}}
\def\twoqtwothree{{\scriptstyle 2(q_2+q_3)}}
\begin{equation}
\setlength{\jot}{2pt}
    \begin{aligned}
&\!\!\!\! \beta(q_2,q_3)\!\!=\!\!
\Bigl[ 1,\s_2,\c_2,\s_3,\c_3,\s_{23},\c_{23},\sin(\twoqtwo),\cos(\twoqtwo),\\
& \quad \sin(\twoqthree),\cos(\twoqthree),     
\sin(\twoqtwothree),\cos(\twoqtwothree))\Bigr]\tr
%         \begin{Bmatrix}
% sin(nq_2),cos(nq_2),sin(nq_3), \\ cos(nq_3),sin(n(q_2+q_3)),cos(n(q_2+q_3)) 
%     \end{Bmatrix}
    % \{ \mathbf{1},\sin(nq_2),\cos(nq_2),\sin(nq_3) ,\cos(nq_3), \\ \sin(n(q_2+q_3)),\cos(n(q_2+q_3) \quad | \quad n\in \{1,2\} \}
    \end{aligned}
\end{equation}
where $\s_i=\sin(q_i)$, $\c_i=\cos(q_i)$, $\s_{ij}=\sin(q_i+q_j)$, $\c_{ij}=\cos(q_i+q_j)$, 
and $\beta\in\mathbb{R}^{13}$.
% \begin{equation}
%     \Theta_\PH = A \phi(q_2,q_3).
%     % \theta_\PH(q) = \mathcal{F}(\theta^n_\PH,A \phi(q_2,q_3))
%     \label{eq:fourierbasis}
% \end{equation}

\vspace{-.0in}
To further reduce the dimension of the basis function, we perform singular value decomposition on the coefficient matrix, $A=U \Sigma V\tr$, and retain $r$ ($r<m$) dominant singular values.  
We then have a reduced basis set: 
%In that case, the original $m$ basis functions in $\phi(q_2,q_3)$ is reduced to $r$ basis functions 
\begin{equation}
\beta_r(q_2,q_3):=
    \begin{bmatrix}
        V_1\tr \\\vdots \\ V_r\tr 
    \end{bmatrix}
    \beta(q_2,q_3).
    \label{eq:beta_r}
\end{equation}
The $A$ matrix may be replaced by a smaller matrix 
\begin{equation}
A_r :=
\begin{bmatrix}
    \sigma_1 U_1 & \ldots & \sigma_r U_r 
\end{bmatrix}.
\label{eq:reducedA}
\end{equation}
The POE parameters in \eqref{eq:fq2q3} may be approximated by
\begin{equation}
    \Theta_{\PH}= A_r \beta_r(q_2,q_3).
\end{equation}

%In this paper, we choose the Fourier basis for $\phi$:

\ifNotEricTrim{
\begin{figure}[hbtp]
    \centering    \includegraphics[width=0.47\textwidth]{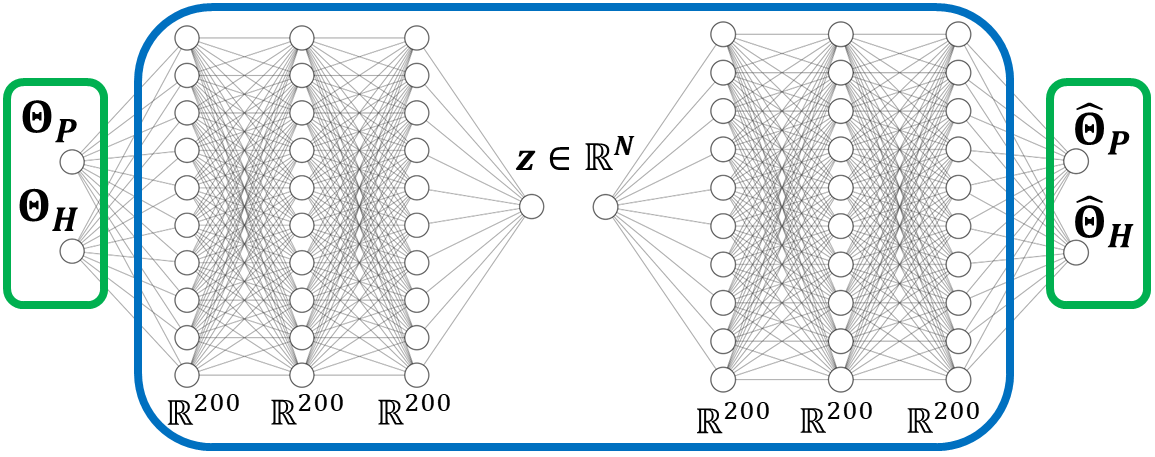}
    \caption{Autoencoder \ifresubmissionblue{\color{blue}\fi (AE) structure\ifresubmissionblue}\fi. The encoder has the parameters $\ERTH$ as the input and latent vector $z$ as the output with the decoder vice versa. They both has three hidden layers with 200 nodes and ReLU as the activation function.}
    \label{fig:ae_structure}
\end{figure}
}\fi

\subsection{Low Dimension Representation with Autoencoder}
\label{sec:cdc_ae}

Given a basis set $\beta$, there may be a lower dimensional basis that more compactly represent the parameter function $f_{\ERTH}$.   The singular value decomposition approach in \eqref{eq:beta_r} finds a lower dimensional hyperplane where the training data are concentrated.  
% % We show in the previous section that the Fourier basis are effective, and lower dimensional representations of the error parameters $\ERTH$. The representations are hand-chosen and inspired by the gravity load symbolic expression. 
% We want learn low-dimensional representations rather than relying on a predefined Fourier basis. 
A more general approach, without the linearity assumption inherent in SVD, is to apply the Autoencoder (AE) method. 
% %However, PCA is merely an linear operation while AE is much powerful with its nonlinearity and can be seen as generalized PCA. 
% We train a set of encoder and decoder. 
The encoder transforms the parameters $\ERTH$ to a low dimension latent vector $z \in \mathbb{R}^N$ as the representation of the original high dimensional data. %Since $\ERTH$ is a function of joint angles $q$, the latent vector $z$ is also a function of $q$. 
The decoder lifts $z$ to an approximate $\ERTH$. Denote the encoder as $f_{en}$, decoder as $f_{de}$ and the decoded parameter as $\hat \Theta_{\PH}$:
\begin{equation}
    z = f_{en}(\ERTH), \quad 
    \hat\Theta_{\PH} = f_{de}(z).
\end{equation}
%The autoencoder learns $f_{en}$ and $f_{de}$ based on a set of training data to minimize $\norm{\ERTH-\hat\Theta_{\PH}}$.
The encoder and decoder are parametrized using neural networks and have similar but flipped network structure\ifNotEricTrim{, visualized in Fig.~\ref{fig:ae_structure}}\fi. 
These networks are trained based on the loss function
\begin{equation}
    \mathcal{L} = \frac{1}{L}\sum_{i=1}^{L}(\ERTH^{(i)}-f_{de}(f_{en}(\ERTH^{(i)})))^2.
\end{equation}

\ifNotEricTrim{
\begin{figure}[ht]
    \centering
    \includegraphics[width=0.35\textwidth]{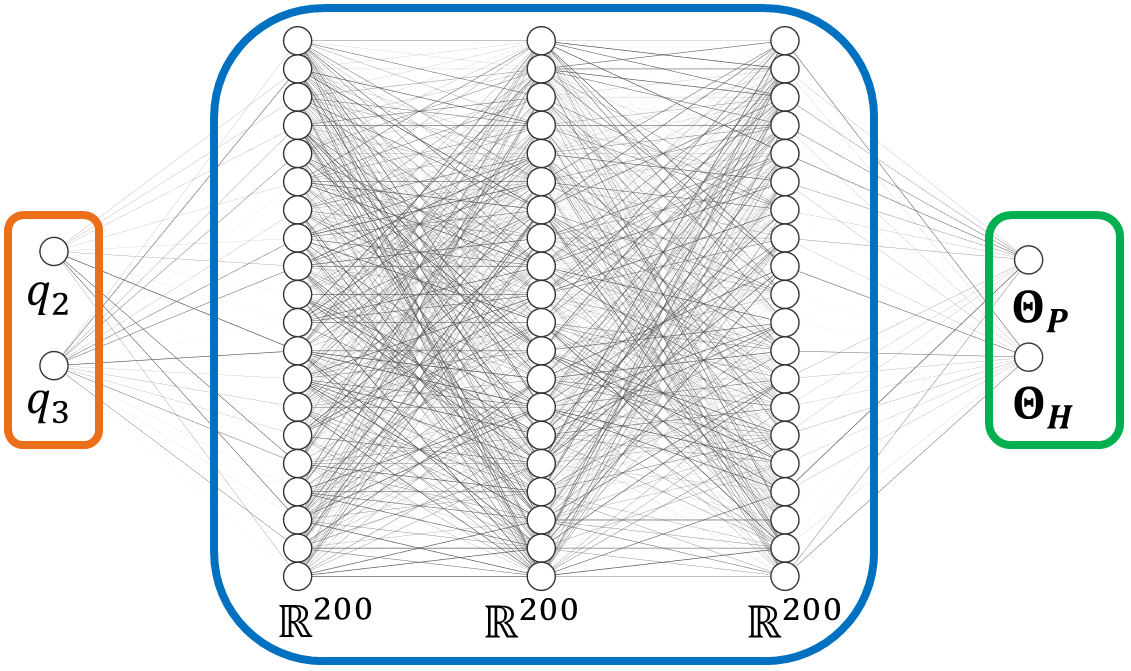}
    \caption{Neural Network (NN) structure. The learned NN approximating the function $f_{\ERTH}(q_2,q_3)$ has one input layer (orange) taking the input of joint 2 $q_2$ and 3 $q_3$, three hidden layers (blue) with 200 nodes each and one output layer (green) predicting the error parameters $\ERTH$.}
    \label{fig:nn_structure}
\end{figure}}\fi

\subsection{Neural Network Function Approximation}
\label{sec:cdc_NN}

We may use a multilayer neural network to directly approximate $f_{\ERTH}(q_2,q_3)$ in \eqref{eq:fq2q3}.
%NN is known to be a universal function approximation, and has been shown effectiveness in robotics context and predicting the residual errors after calibration. Here we would like to directly learn the kinematic parameters using NN.
\ifNotEricTrim{The structure of the NN is showed in Fig.~\ref{fig:nn_structure}.}\fi  
\ifEricTrim{Here is our NN structure. }\fi
The input layer has two nodes, i.e. the joint 2 and 3 angles. The hidden layers are fully connected layers with ReLU activation functions. The output layers is a fully connected layer with no output function, which become the linear combination of the hidden layers output. 
%Supposed $\ERTH$ parameters are obtained for $L$ clusters, 
The loss function is the mean square error between the predicted parameters and the calibrated ones.
\begin{equation}
    \mathcal{L} = \frac{1}{L}\sum_{i=1}^{L}(\hat \Theta_{\PH}(q^{(i)})-\Theta_{\PH}(q^{(i)}))^2
\end{equation}
where $\hat \Theta_{\PH}(q^{(i)})$ is the predicted output from the NN. 
%Besides the loss function, we further evaluate the mean and max position error using the estimated $\ERTH$ from the NN. 
%so that we know how the NN actually performs in estimating the true parameters.
%Since there are relatively small numbers of data samples, 
%The training batch size equals to the number of data samples, i.e. one batch only. 
We use the Adam optimizer \cite{Kingma2014AdamAM} to update the parameters  based on the loss of each learning epoch with adaptive learning rate.  
\section{Implementation and Evaluation}
\label{sec:exp}

\subsection{Experiment Setup}

\begin{figure}[ht]
     \centering
     \begin{tabular}{cc}
%     \begin{subfigure}[b]{0.25\textwidth}
%         \centering
         \includegraphics[width=.23\textwidth]{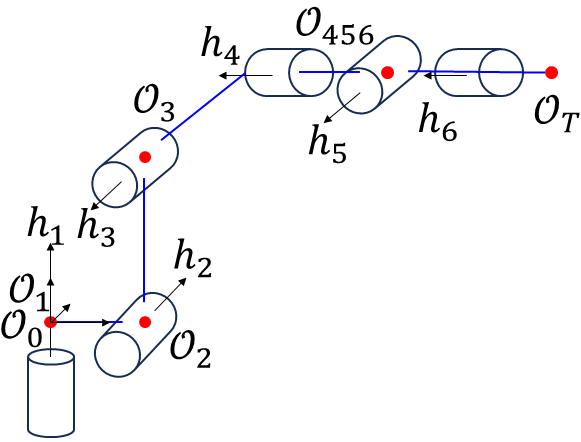}
%         \caption{MA2010 Schematics} 
%         \label{fig:r1_scheme}
%     \end{subfigure}
%     \begin{subfigure}[b]{0.22\textwidth}
%         \centering
 &        \includegraphics[width=.21\textwidth]{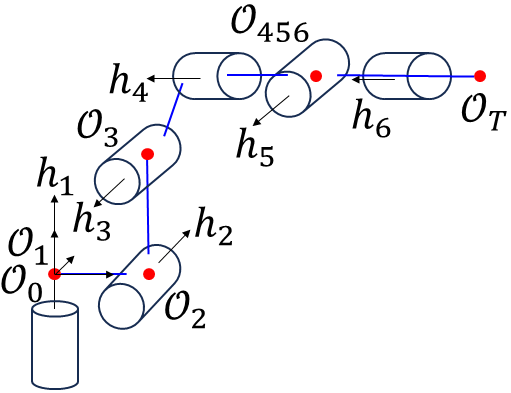}
%         \caption{MA1440 Schematics}
%         \label{fig:r2_scheme}
%     \end{subfigure}
\\
${\bf P}  = \begin{bmatrix}
    \begin{matrix} 0 \\ 0 \\ 0 \end{matrix}\,\,
    \begin{matrix} 150 \\ 0 \\ 0 \end{matrix}
    \begin{matrix} 0 \\ 0 \\ 760 \end{matrix}\,
    \begin{matrix} 1082 \\ 0 \\ 200 \end{matrix}\,\,
    \begin{matrix} 0 \\ 0 \\ 0 \end{matrix}\,\,
    \begin{matrix} 100 \\ 0 \\ 0 \end{matrix}    
\end{bmatrix}$
&
${\bf P}  = \begin{bmatrix}
    \begin{matrix} 0 \\ 0 \\ 0 \end{matrix}\,\,
    \begin{matrix} 155 \\ 0 \\ 0 \end{matrix}
    \begin{matrix} 0 \\ 0 \\ 614 \end{matrix}\,
    \begin{matrix} 640 \\ 0 \\ 200 \end{matrix}\,\,
    \begin{matrix} 0 \\ 0 \\ 0 \end{matrix}\,\,
    \begin{matrix} 100 \\ 0 \\ 0 \end{matrix}    
\end{bmatrix}$\\
${\bf H} = \begin{bmatrix}
    \begin{matrix} 0 \\ 0 \\ 1 \end{matrix}\,\,
    \begin{matrix} 0 \\ 1 \\ 0 \end{matrix}\,\,
    \begin{matrix} 0 \\ -1 \\ 0 \end{matrix}\,\,
    \begin{matrix} -1 \\ 0 \\ 0 \end{matrix}\,\,
    \begin{matrix} 0 \\ -1 \\ 0 \end{matrix}\,\,
    \begin{matrix} -1 \\ 0 \\ 0 \end{matrix}    
\end{bmatrix}$
&
${\bf H} = \begin{bmatrix}
    \begin{matrix} 0 \\ 0 \\ 1 \end{matrix}\,\,
    \begin{matrix} 0 \\ 1 \\ 0 \end{matrix}\,\,
    \begin{matrix} 0 \\ -1 \\ 0 \end{matrix}\,\,
    \begin{matrix} -1 \\ 0 \\ 0 \end{matrix}\,\,
    \begin{matrix} 0 \\ -1 \\ 0 \end{matrix}\,\,
    \begin{matrix} -1 \\ 0 \\ 0 \end{matrix}    
\end{bmatrix}$
\\
\\
(a) MA2010 (R1) & (b) MA1440 (R2)
\end{tabular}
    \caption{Robot schematics in the zero configuration.}
    \label{fig:robot_scheme}
\end{figure}

For the experimental verification of the calibration methods, we used Yaskawa MA2010 and MA1440 robots from our welding testbed. Both robots are the classic 6 rotational-axis (6R) industrial robot elbow arm with spherical wrists (joints 4, 5, and 6 have intersecting rotation axes). 
% The robot's kinematic diagram is in Fig. \ref{fig:robot_diag}, and the ideal POE model parameters are presented in Table \ref{tab:nominal_ma2010}..
The vendor-provided POE model parameters and the zero configurations of these robots are shown in Fig.~\ref{fig:robot_scheme}. %\eqref{eq:POErobots}.

% \begin{table}[htbp]
% \begin{center}
% \begin{tabular}{c c c c} 
% \toprule
% Link Vectors & Parameters & Rotation Axis & Parameters \\
% \midrule
% P1 & [0,0,0] & H1 & [0,0,1] \\ 
% P2 & [150,0,0] & H2 & [0,1,0]\\
% P3 & [0,0,760] & H3 & [0,-1,0] \\
% P4 & [1082,0,200] & H4 & [-1,0,0] \\ 
% P5 & [0,0,0] & H5 & [0,-1,0] \\
% P6 & [0,0,0] & H7 & [-1,0,0] \\
% P7 & [100,0,0] & &\\
% \bottomrule
% \end{tabular}
% \caption{Nominal POE Kinematic Model Parameters of MA2010 from Yaskawa Motoman}
% \label{tab:nominal_ma2010}
% \end{center}
% \end{table}

% \begin{table}[htbp]
% \begin{center}
% \begin{tabular}{c c c c} 
% \toprule
% Link Vectors & Parameters & Rotation Axis & Parameters \\
% \midrule
% P1 & [0,0,0] & H1 & [0,0,1] \\ 
% P2 & [155,0,0] & H2 & [0,1,0]\\
% P3 & [0,0,614] & H3 & [0,-1,0] \\
% P4 & [640,0,200] & H4 & [-1,0,0] \\ 
% P5 & [0,0,0] & H5 & [0,-1,0] \\
% P6 & [0,0,0] & H7 & [-1,0,0] \\
% P7 & [100,0,0] & &\\
% \bottomrule
% \end{tabular}
% \caption{Nominal POE Kinematic Model Parameters of MA1440 from Yaskawa Motoman}
% \label{tab:nominal_ma1440}
% \end{center}
% \end{table}

To measure the tool pose, $T_{0T}$, we use the OptiTrack motion capture system, with eight Prime\textsuperscript{x}13 \cite{primex13} and the software Motive.
% {\color{red} model type, citation}. 
\ifNotEricTrim{This system operates by detecting retro-reflective markers using IR-emitting cameras
%. The calibrated cameras capture reflected IR from the markers 
to estimate the distance to each marker.}\fi The stated accuracy for the marker distance is $0.2$ mm. Using Small-Angle approximation, the orientation accuracy is approximately $\theta \approx \frac{0.2}{L}$, where $L$ is the distance between the markers. In our setup, five markers are attached to the end effector of each robot. The orientation accuracy is around $0.05$ degree. \ifresubmissionblue{\color{blue}\fi The 2D camera resolution is about 0.02-0.03 mm on a 2D plane enable by the sub-pixel centroiding technology. \ifresubmissionblue}\fi
%and is capable of tracking multiple markers simultaneously. 
%Furthermore, the system allows users to group a set of markers as a "rigid body," enabling the creation of a coordinate frame from the detected markers. 
The setup of the robot and the motion tracking system is shown in Fig.~\ref{fig:optitrack_setup}.  
% The motion tracking system has a total of 8 cameras.
\ifresubmissionblue{\color{blue}\fi We collect data only after the system is properly calibrated and has reached thermal stability to ensure optimal accuracy.\ifresubmissionblue}\fi

For each robot, we collect a training set of tool frame data, $T_{0T}$, and the corresponding joint angles, $q$,
and a test set for evaluation of the calibration outside of the training set. 
%The motion capture system is used for all the tool frame data collection.
%
For MA2010, we specify 248  configurations with $q_2\in[-55^\circ,50^\circ]$ and  $q_3\in[-70^\circ,50^\circ]$ as anchors for the parameter identification clusters. 
%The exact sampled configurations are shown in Fig. \ref{fig:ph_q2q3}. 
Around each anchor configuration, we collect 7 sets of tool frame data for a total of 1736 tool frame data points in 248 clusters in the training set. 
For MA1440, the training data consists of 280 configurations with $q_2\in [-55^\circ, 50^\circ]$ and $q_3\in[-70^\circ,60^\circ]$ as cluster anchors for a total of 1960 tool frame data points in 280 clusters in the training set.

\subsection{Evaluation}

We apply the NLS and CDC methods on the training data to find the kinematics model and evaluate the result on the same training dataset. 
The NLS method finds one set of $(\PH)$ to minimize all the tool errors. The CDC method finds one set of $(\PH)$ for each cluster. \ifresubmissionblue{\color{blue}\fi We include an additional calibration case in our evaluation: a base-only calibration (Base), where only the base position (3 parameters) is optimized. This scenario corresponds to redefining the reference frame to eliminate any fixed base offset. \ifresubmissionblue}\fi
In our manufacturing application of cold spraying, the performance  requirement is less than 1~mm in position error and 1~degree in the surface normal error. To balance these two requirements in the objective function \eqref{eq:trans_norm}, we set the weighting factor to $a=1/(8\sin^2(\frac{\pi}{360}))=1641$.
% We choose the weighting factor $a=\frac{180}{\pi}$ in Eq.\ref{eq:trans_norm} because our manufacturing application, cold spraying, requires a sub-millimeter and sub-degree level of accuracy.
The error statistics for the training set is summarized in Table~\ref{tab:training_data_error}. 
% The tool position error 
% % {\color{red} is it full position and orientation error or just position error?} 
% for all training samples is shown in Fig.~\ref{fig:r1_training_error}--\ref{fig:r2_training_error}.
Both NLS and CPA improves on the vendor parameters, but CPA only works well on certain configurations and thus has a large standard deviation. 
%On the other hand, 
Configuration dependency in CDC allows further reduction of the tool frame error over the training set.  
% {\color{red} What about CPA?}
\ifresubmissionblue{\color{blue}\fi The base-only calibration reduces the mean position error but does not achieve the same level of improvement as NLS or CDC. Moreover, the orientation error remains unchanged, indicating that the error is not simply due to a fixed offset. 
\ifresubmissionblue}\fi

\begin{table}[h!]
\centering
\begin{tabular}{c c c c c c c}
\hline
\hline
 &
\multicolumn{3}{c}{R1 Position Error (mm)} & \multicolumn{3}{c}{R2 Position Error (mm)} \\
Method & Mean &  Std & Max & Mean & Std & Max\\
\hline
Nominal & 1.32 & 0.23 & 1.81 & 1.77 & 0.94 & 4.18 \\
Base & 0.41 & 0.20 & 1.24 & 1.59 & 0.89 & 3.49 \\
CPA & 0.49 & 0.25 & 1.20 & 1.59 & 0.79 & 3.29 \\
NLS & 0.25 & 0.14 & 0.86 & 0.41 & 0.28 & 1.53 \\
CDC & \textbf{0.10} & \textbf{0.05} & \textbf{0.46} & \textbf{0.04} & \textbf{0.02} & \textbf{0.12} \\
\hline
\hline
& \multicolumn{3}{c}{R1 Orientation Error (deg)} & \multicolumn{3}{c}{R2 Orientation Error (deg)} \\
Method & Mean &  Std & Max & Mean & Std & Max\\
\hline
Nominal & 0.07 & 0.02 & 0.12 & 0.12 & 0.04 & 0.26 \\
Base & 0.07 & 0.02 & 0.12 & 0.12 & 0.04 & 0.26 \\
CPA & 0.05 & 0.02 & 0.13 & 0.13 & 0.05 & 0.28 \\
NLS & 0.05 & 0.02 & 0.11 & 0.10 & 0.04 & 0.27 \\
CDC & \textbf{0.04} & \textbf{0.02} & \textbf{0.10} & \textbf{0.09} & \textbf{0.05} & \textbf{0.25} \\
\hline
\hline
\end{tabular}
\caption{Tool frame position and orientation error statistics comparison on the training data. \ifresubmissionblue{\color{blue}\fi 
%Asterisk$(^*)$ shows the propose methods and the 
Boldface font highlights the best performance among all methods.
\ifresubmissionblue}\fi  
}
\label{tab:training_data_error}
\end{table}

\begin{figure}[hbtp]
    \centering    
    \includegraphics[width=0.46\textwidth]{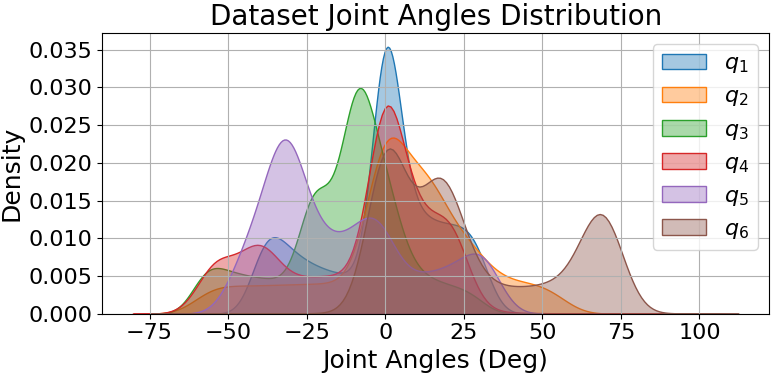}
    \caption{\ifresubmissionblue{\color{blue}\fi Joint angle distribution in the testing dataset shows broad coverage of all joint angles.
    %, ure comprehensive coverage and improve the representativeness of the dataset.
\ifresubmissionblue}\fi}
    \label{fig:data_joint_dist}
\end{figure}

To evaluate the performance of the identified POE models beyond the training set,
%we compare the tool position error using the testing dataset.
\ifresubmissionblue{\color{blue}\fi 
we use a testing dataset consisting of 1,500 robot poses, uniformly sampled from each robot's workspace. 
The testing dataset is representative of the entire workspace, as shown by the joint angle distribution in Fig.~\ref{fig:data_joint_dist}.
%from random arm configurations of each robot in the workspace to assess the performance of the calibration methods. 
\ifresubmissionblue}\fi
\ifresubmissionblue{\color{blue}\fi 
The error from applying the nominal POE model to the testing dataset confirms the strong correlation between the position error in the clusters and the shoulder and elbow angles, as shown in Table~\ref{tab:error_joint_cc}. 
%This indicates that the identified parameters $\ERTH^{(\ell)}$ in the clusters correlate most significantly with $(q_2, q_3)$.

%We examined our dataset to verify this assumption. The dataset was collected to ensure sufficient angle coverage for all joints ($q_1$ to $q_6$), making as much representative of the full joint space in our working cell. The distribution of the collected joint data is shown in Fig.~\ref{fig:data_joint_dist}. Table~\ref{tab:error_joint_cc} shows the correlation between position error and joint angles. We found that the shoulder and elbow joint angles exhibit the highest correlation, indicating that the parameters $\ERTH^{(\ell)}$ change most significantly with variations in $(q_2, q_3)$. 
\ifresubmissionblue}\fi

\begin{table}[h!]
\centering
\begin{tabular}{c c c c c c c }
\hline
\hline
 & $q_1$ & $q_2$ & $q_3$ & $q_4$ & $q_5$ & $q_6$ \\
\hline
MA2010 (R1) & {0.16} & {\bf 0.83} & {\bf 0.82} & 0.30 & 0.61 & 0.31 \\
MA1440 (R2) & 0.38 & {\bf 0.77} & {\bf 0.44} & 0.31 & 0.35 & 0.07 \\
\hline
\end{tabular}
\caption{\ifresubmissionblue{\color{blue}\fi Correlation of the position errors with the joint angles show the strongest correlation with the shoulder and elbow joint angles ($q_2,q_3$).
%The highest correlation between shoulder and elbow joint angles ($q_2,q_3$) and positions errors indicate the parameters  most significantly with variations in $q_2,q_3$.
\ifresubmissionblue}\fi}
\label{tab:error_joint_cc}
\end{table}
%calibrated parameters with the measured data using the testing dataset.  We compare the following schemes:

\ifresubmissionblue{\color{blue}\fi 
We consider the following POE parameter identification methods for tool pose error comparison on the testing dataset:
\ifresubmissionblue}\fi
\begin{itemize}
\setlength{\leftskip}{-1em}
\item Non-Configuration-Dependent (one set of $(\PH)$ parameters for all configurations)
\begin{itemize}
\setlength{\leftskip}{-2em}
    \item Nominal $(\PH)$: Vendor-provided  parameters.
    \item \ifresubmissionblue{\color{blue}\fi Base: Optimization of the robot base frame position, corresponding to a 3-parameter kinematic calibration. \ifresubmissionblue}\fi
    \item CPA: Circular point analysis method (Secton~\ref{sec:cpa}) around the zero configuration, $q=0$.
    %, as in Section~\ref{sec:cpa}.
    \item NLS-0: Nonlinear least square method (Section~\ref{sec:NLS}) around the zero configuration, $q=0$.
    \item NLS-1: Nonlinear least square method based on all training data.
\end{itemize}
\item Configuration-Dependent Kinematics based on the interpolation from the NLS fit for each training cluster.
\begin{itemize}
\setlength{\leftskip}{-2em}
    \item Nearest Neighbor: Nearest cluster.
    \item Linear Interpolation: Linear interpolation based on nearby clusters.
    \item Cubic Interpolation: Cubic spline interpolation based on nearby clusters.
    \item RBF Interpolation: Radial basis function based on the training data (Section~\ref{sec:cdc_fourier}) 
    \item FBF Interpolation: Fourier basis function based on the training data (Section~\ref{sec:cdc_fourier}) The basis dimension $N=13$ and $N=7$ (reduced basis).
    % {\color{red} Did you use the reduced basis?  What is the basis dimension?}
    \item NN Interpolation: Neural Network approximation based on the training data (Section~\ref{sec:cdc_NN})
    \item AE Interpolation: Linear interpolation of the latent space vectors and decoding to $\ERTH$ (Section~\ref{sec:cdc_ae}). Latent space dimension $N=6$.
    % {\color{red} What is $N$ that you used here?}
\end{itemize}
\end{itemize}
\ifresubmissionblue{\color{blue}\fi 
For FBF, NN and AE, the number of parameters for each method is given in Table~\ref{tab:num_of_params}.
\ifresubmissionblue}\fi
\begin{table}[h!]
\centering
\begin{tabular}{c c c c c c}
\hline
\hline
Robots & FBF N=13 & FBF N=7 & NN & AE N=6 \\
\hline
R1 & 312 & 168 & 87633 & 88420 \\
R2 & 312 & 168 & 87633 & 88420 \\
\hline
\hline
\end{tabular}
\caption{\ifresubmissionblue{\color{blue}\fi Number of parameters used in each method. For AE, only the number of parameters in the decoder, which used for estimation, is reported. \ifresubmissionblue}\fi}
\label{tab:num_of_params}
\end{table}
\smallskip
\begin{table}[htbp]
\centering
\begin{tabular}{c c c c c c c}
\hline
\hline
&\multicolumn{3}{c}{R1 Position Error (mm)} & \multicolumn{3}{c}{R2 Position Error (mm)} \\
Method & Mean & Std & Max & Mean & Std & Max\\
\hline
Nominal & 1.29 & 0.21 & 1.72 & 1.08 & 0.81 & 3.07\\
Base & 0.53 & 0.25 & 1.27 & 1.17 & 0.62 & 2.93\\
CPA & 0.44 & 0.20 & 0.99  & 1.44 & 0.61 & 3.19 \\
NLS-0 & 0.41 & 0.21 & 0.99 & 1.68 & 0.71 & 3.20\\
NLS-1 & 0.92 & 0.79 & 3.46 & 1.71 & 1.20 & 5.18\\
Nearest & 0.32 & 0.17 & 0.81 & 0.46 & 0.29 & 1.29\\
Linear & 0.31 & 0.17 & 0.81 & 0.42 & 0.26 & 1.23\\
Cubic  & 0.31 & 0.17 & 0.83 & 0.43 & 0.27 & 1.33\\
RBF & 0.31 & 0.17 & 0.83 & 0.47 & 0.44 & 4.90\\
FBF N=13 & 0.31 & 0.14 & 0.74 & 0.42 & 0.21 & 0.99\\
FBF N=7 & 0.32 & 0.16 & 0.81 & 0.44 & 0.19 & 0.98\\
NN & 0.34 & \textbf{0.13} & 0.78 & 0.43 & \textbf{0.15} & \textbf{0.77}\\
AE N=6 & \textbf{0.31} & 0.14 & \textbf{0.73} & \textbf{0.42} & 0.19 & 0.94\\
\hline
\hline
&\multicolumn{3}{c}{R1 Orientation Error (deg)} & \multicolumn{3}{c}{R2 Orientation Error (deg)} \\
Method & Mean & Std & Max & Mean & Std & Max\\
\hline
Nominal & 0.10 & 0.04 & 0.20 & 0.11 & 0.04 & 0.23\\
Base & 0.10 & 0.04 & 0.20 & 0.11 & 0.04 & 0.23\\
CPA & \textbf{0.08} & \textbf{0.03} & 0.17 & 0.11 & 0.04 & 0.20\\
NLS-0 & 0.08 & 0.03 & \textbf{0.17} & 0.11 & 0.05 & 0.22\\
NLS-1 & 0.19 & 0.09 & 0.39 & 1.00 & 0.58 & 2.24\\
Nearest & 0.09 & 0.04 & 0.19 & 0.09 & 0.04 & 0.19\\
Linear & 0.09 & 0.04 & 0.19 & 0.09 & 0.04 & 0.19\\
Cubic  & 0.09 & 0.04 & 0.19 & 0.09 & 0.04 & 0.19\\
RBF & 0.09 & 0.04 & 0.19 & 0.09 & 0.04 & 0.31\\
FBF N=13 & 0.09 & 0.04 & 0.18 & \textbf{0.09} & 0.04 & 0.20\\
FBF N=7 & 0.09 & 0.04 & 0.18 & 0.09 & \textbf{0.04} & \textbf{0.19}\\
NN & 0.10 & 0.04 & 0.19 & 0.09 & 0.04 & 0.20\\
AE N=6 & 0.09 & 0.04 & 0.19 & 0.09 & 0.04 & 0.21\\
\hline
\hline
\end{tabular}
\caption{Tool frame position and orientation error of multiple calibration methods applied to the test data. \ifresubmissionblue{\color{blue}\fi 
%Asterisks ($*$) indicate the proposed methods, and 
Boldface font highlights the best performance among all methods.
\ifresubmissionblue}\fi}
\label{tab:testing_data_error}
\end{table}

\begin{figure*}[hbtp]
    \centering
    \begin{subfigure}[b]{0.45\textwidth}
        \centering
        \includegraphics[width=\textwidth]{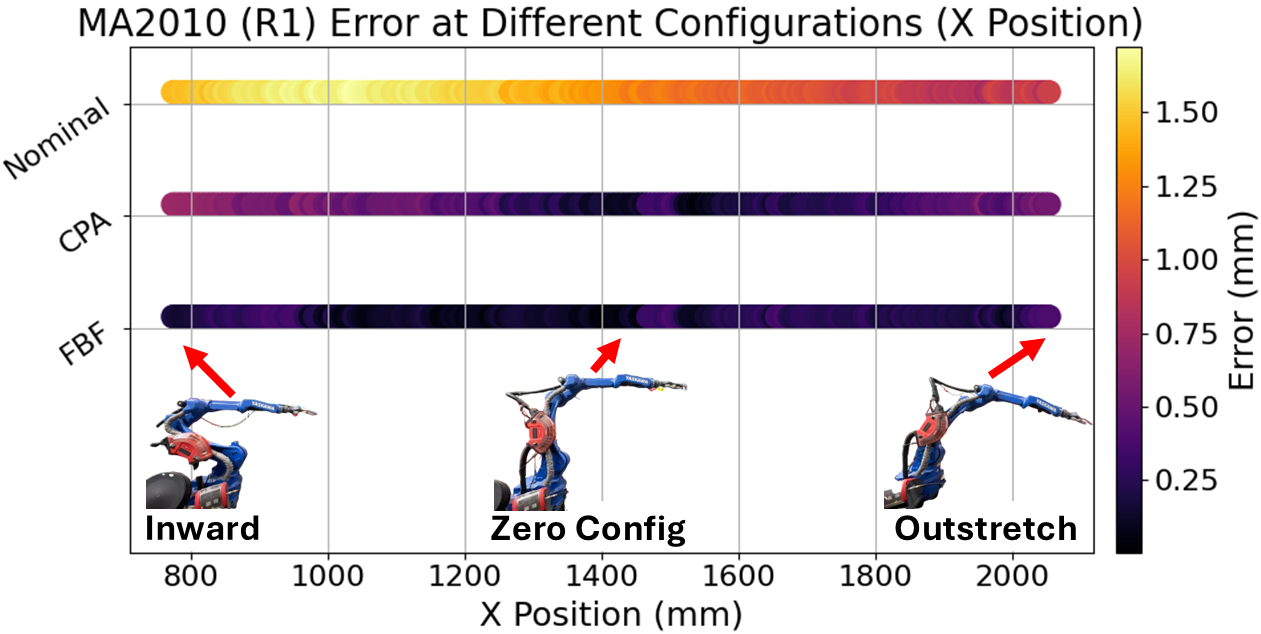}
        \caption{MA2010 testing dataset position error.}
        \label{fig:r1_testing_error}
    \end{subfigure}
    \begin{subfigure}[b]{0.45\textwidth}
        \centering
        \includegraphics[width=\textwidth]{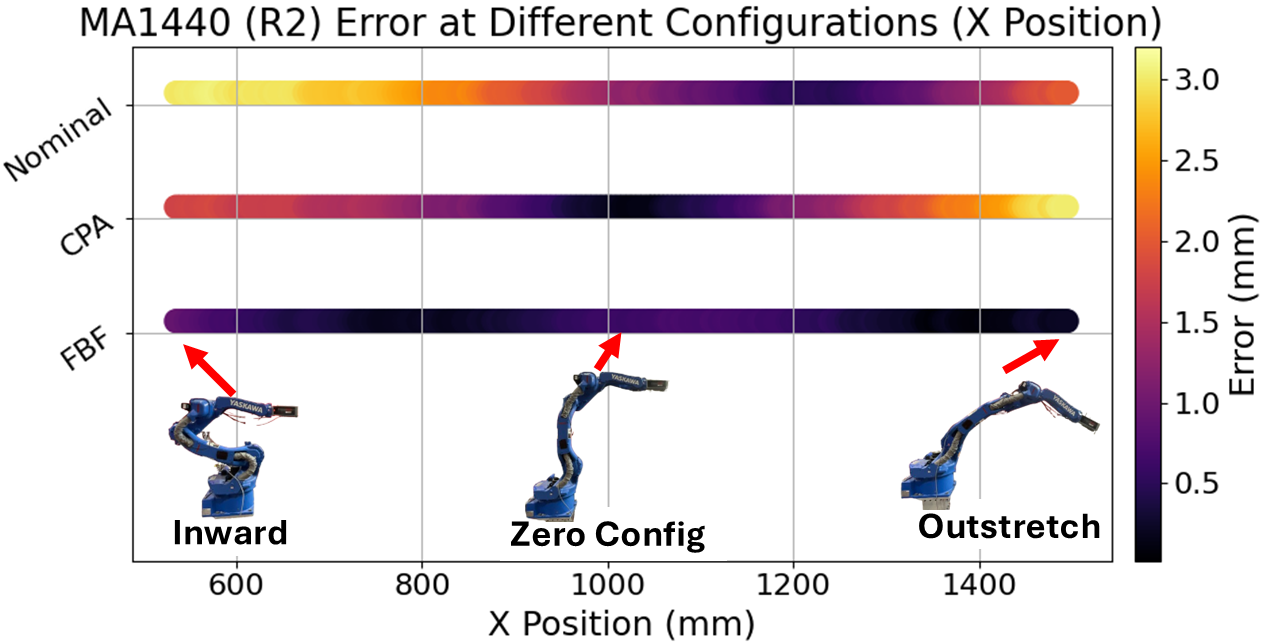}
        \caption{MA1440 testing dataset position
        error.}
        \label{fig:r2_testing_error}
    \end{subfigure}
    \caption{R1 (MA2010), R2 (MA1440) testing dataset position error. \ifresubmissionblue{\color{blue}\fi The $X$-Position represents the tool approach direction in the  robot base frame, which corresponds to different robot configurations: from inward (left), through zero configuration (center), to outstretched (right). The color bars correspond to the three POE parameter sets.  
    % The colormap indicates the error magnitude in millimeters. 
    For the nominal parameters and circular point analysis (CPA) method, the tool pose error varies with configuration. CPA calibrated at the zero configuration shows the lowest error near its calibration pose, but larger error at other configurations. Configuration-dependent methods such as the Fourier basis function (FBF) method maintains consistently low error across the entire workspace.  R2 also shows higher error variation across configurations, leading to poor performance of standard interpolation methods. \vspace{-7mm}
    \ifresubmissionblue}\fi}
\end{figure*}

At each configuration in the testing dataset, we compare the computed tool pose using the identified POE parameters with the measured tool pose. 
Fig.~\ref{fig:r1_testing_error}-\ref{fig:r2_testing_error} show the tool frame position error of the methods applied to the two robots \ifresubmissionblue{\color{blue}\fi 
%at different configurations 
in each robot's base frame. \ifresubmissionblue}\fi Table~\ref{tab:testing_data_error} summarizes the statistics of the tool frame position error for the different parameter identification methods.   
% We conducted both paired t-tests and Wilcoxon signed-rank tests on the position error data between Configuration-Dependent Kinematics methods and Non-Configuration-Dependent ones. In all cases, the p-values were extremely close to zero, indicating that the differences in performance are statistically significant and our method provides a meaningful improvement over existing approaches.
\ifresubmissionblue{\color{blue}\fi
Paired t-tests and Wilcoxon signed-rank tests  comparing configuration-dependent and non-configuration-dependent methods all show p-values near zero, indicating statistically significant improvements.
\ifresubmissionblue}\fi 
% \ifresubmissionblue{\color{blue}\fi 
% We showed the calibrated $\ERTH$, mean and variation across all configurations in Table~\ref{tab:PH_var}.
% \ifresubmissionblue}\fi
From these results, we make the following observations:

\begin{itemize}
\setlength{\leftskip}{-1em}
%
% improvement of position, less so on orientation
% 
% interpolation works better if ther variation between cluster is small (like for R1).  When the variation is larger (like R2), smoothing using basis functions can improve interpolation accuracy. 
%
    \item All calibration methods, except for NLS with fixed POE parameters,  tend to improve on the nominal vendor-provided parameters. This highlights the importance of calibration.
    \item \ifresubmissionblue{\color{blue}\fi
    For R1, FBF and NN methods do not improve much over standard interpolation methods.  For R2, the difference is more pronounced. This may be explained by examining the variation of locally identified POE parameters $\ERTH$ from the nominal values over the training dataset, as shown in Table~\ref{tab:PH_var}. 
    The higher standard deviation in R2 indicates that it has much more variations across the configurations than R1.  This implies that standard interpolation methods would generalize poorly than FBF and NN methods. 
\begin{table}[h!]
\centering
\begin{tabular}{c c c c c}
\hline
\hline
& \multicolumn{2}{c}{$\ERP$ (mm)} & \multicolumn{2}{c}{$\ERH$ (deg)} \\
Robots &  Mean & Std & Mean & Std\\
\hline
R1 & 0.6653 & 0.0412 & 0.0353 & 0.0131\\
R2 & 0.6070 & 0.1392 & 0.1131 & 0.0429\\
\hline
\hline
\end{tabular}
\caption{\ifresubmissionblue{\color{blue}\fi 
Statistics of the deviation locally identified POE parameters from the nominal POE parameters over the training dataset. R2 shows much higher POE parameter variation between configuration.
%The mean and standard deviation of POE parameters at all configuration. R2 exhibits a higher discrepancy between the nominal and actual kinematic and greater $\ERTH$ variation across configurations. This contributes to the higher position error and error variation observed in R2.
\ifresubmissionblue}\fi}
\label{tab:PH_var}
\end{table}
    
    %    Table~\ref{tab:PH_var} lists the statistics of the variation of locally identified 
%    $\ERTH$ from the nominal values over the training dataset.
%    calibrated $\ERTH$ mean and standard deviation across all configurations. 
%It is clear that R2 exhibits a larger offset from the nominal, indicating greater discrepancy between its nominal and actual models, which contributes to its higher position error. R2 also shows greater $\ERTH$ variation across configurations, leading to higher error variation when using CPA or Nominal. 
\ifresubmissionblue}\fi
    \item When the variation of $\ERTH$ is small between clusters (i.e., low configuration dependence), all interpolation methods perform reasonably well (as in the R1 case in Table~\ref{tab:testing_data_error}).  When the variation is large, interpolation using basis functions performs significantly better (as in the R2 case in Table~\ref{tab:testing_data_error}).   
%    With a larger configuration dependency (e.g., R2 to R1 in this case), Fourier basis, NN, and AE interpolation yield better performance 
%{\color{red}put PH variation figure}.
    \item CPA works the best near the configuration where the process is applied.  It also works reasonably well in other configurations but with degraded performance away from the nominal configuration (the zero configuration in our case).
    \item The NLS method with fixed POE parameters, whether using data around the zero configuration or all data from the workspace, works well on the training set (as in Table~\ref{tab:training_data_error}) but does not generalize well away from the training set.    
    \item CDC methods work well at the training data but the interpolation method affects its performance at the test configurations. This is most evident for the RBF interpolation method where large error could result between the training points. 
    \item With the universal function approximation capability, the NN interpolation achieves the best accuracy among all interpolation methods, though the training process is time consuming. 
    \item The Fourier basis interpolation motivated from the gravity load works comparably to NN interpolation. The chosen basis function has likely captured the functional dependence of the kinematics on configurations. Using only the dominant basis (7), we can reduce the number of basis functions by almost half, without sacrificing much performance. \ifresubmissionblue{\color{blue}\fi In addition, FBF requires the fewest parameters to approximate the kinematic model, resulting in greater training efficiency. \ifresubmissionblue}\fi
    \item The autoencoder method reduces the dimension of the basis vectors (latent space) and still achieves significant improvement of the tool pose prediction error.  The dimension of the latent space (6) is about the same as the same as the dimension of the reduced Fourier basis (7).
\end{itemize}

\noindent{{\em a)} \bf Fourier Basis Reduction}
%Using MA2010 robot as the example, we collect the training data to identify $(\PH)$/$\ERTH$ parameters at 248 robot configurations.  
%We then use the Fourier basis set to estimate the matrix $A$. 
% {\color{red} Did you include this in the performance comparison?}
\ifNotEricTrim{
For MA2010 (R1), 
Fig.~\ref{fig:coeff_ana} shows the SVD results of $A$. The left plot shows the log-scale of the singular values of $A$. Approximately seven singular values appear to be significant. The right plot shows a heat map of the magnitude of the $V^T$ matrix.  The first few rows contain the constant function basis $\mathbf{1}$, as well as $\cos(q_2)$ and $\cos(q_3)$ with larger values. This suggests that these basis functions have a more significant influence on the configuration-dependent kinematic models. 
%As new data is added to the training samples, one can employ updates to adjust the reduced coefficient matrix $A$ in \eqref{eq:reducedA}. 
As shown in Table \ref{tab:testing_data_error},
the performance of the reduced basis is  comparable to using the full basis.
}\fi
\ifEricTrim{
We do SVD for the coefficient matrix $A$ of MA2010 (R1). There are approximately seven singular values appear to be significant. We also finds that basis $\cos(q_2)$, $\cos(q_3)$ and constant function $\mathbf{1}$ have a more significant influence on the configuration-dependent kinematic models. As shown in Table \ref{tab:testing_data_error}, the performance of the reduced basis is  comparable to using the full basis.
}\fi

\ifNotEricTrim{
\begin{figure}[ht]
     \centering
     \begin{subfigure}[b]{0.26\textwidth}
         \centering
         \includegraphics[width=\textwidth]{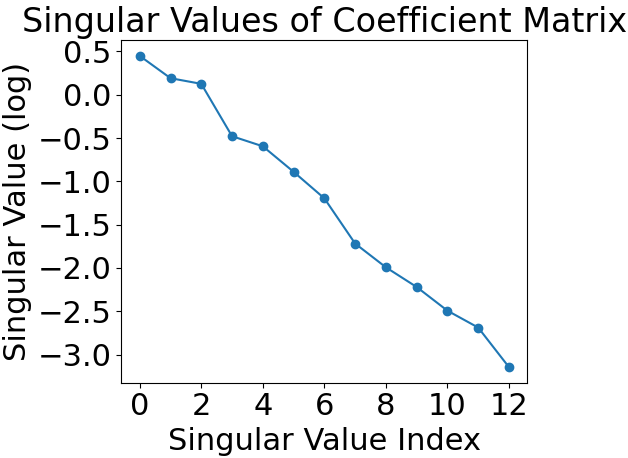}
         \caption{Singular value of Matrix $A$}
         \label{fig:sing_val_A}
     \end{subfigure}
     \begin{subfigure}[b]{0.215\textwidth}
         \centering
         \includegraphics[width=\textwidth]{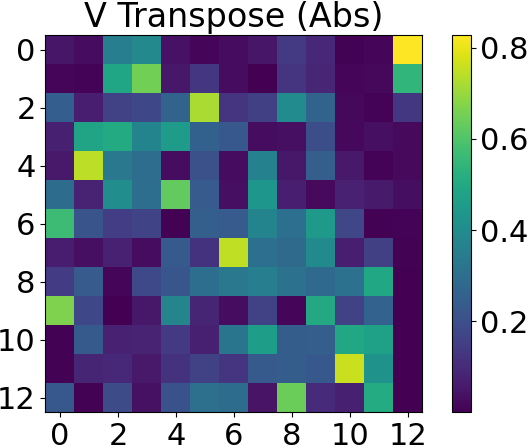}
         \caption{Matrix $V^T$}
         \label{fig:mat_vt}
     \end{subfigure}
    \caption{Coefficient analyzation of the basis function through SVD. On the left is the singular value of matrix $A$ in log scale. On the right is the matrix $V^T$.}
    \label{fig:coeff_ana}
\end{figure}
}\fi

\noindent{{\em b) \bf Autoencoder Implementation}
%The encoder has the parameters $\ERTH$ as the input and latent vector $z$ as the output with the decoder vice versa. 
We implement both the encoder and decoder with three hidden layers.  Each layer has 200 nodes and ReLU as the activation function with no activation function at the output layer and thus the last layer is a linear function. 
% {\color{red} s \bf what does this mean?}. 
% For training, the loss function is the reconstruction mean square error. 
%We compare the accuracy of different latent space dimension $N$. 
%Because we had 12 Fourier basis plus a bias, we here tested out $N=2,6,12$.  
Given a configuration $q$, we linearly interpolate from the identified POE to obtain the latent vector $z$ and then decode $z$ to obtain the parameter $\ERTH$.
Table~\ref{tab:ae_latent} shows that with dimension $N=6$, the decoded parameters have the best accuracy.  Further examination of the structure of the latent space
indicates a strong relationship between the latent vector distribution and joint 2 and 3. 
\begin{table}[htbp]
\centering
\begin{tabular}{c c c c }
\hline
\hline
$N$ & 2 & 6 & 12 \\
\hline
Mean & 0.33 & 0.34 & 0.32 \\
Max & 0.98 & 0.86 & 0.91 \\
\hline
\hline
\end{tabular}
% \caption{Tool frame position error statistics comparison on the training data, using either numerical (Num) or analytical (ana) gradient matrix. }
\caption{Mean and max testing position error (mm) of different latent space dimension $N$.}
\label{tab:ae_latent}
\end{table}

\noindent{{\em c)} \bf Neural Network Approximation}
\ifresubmissionblue{\color{blue}\fi In \cite{Hsiao2020positioning}, the authors used neural networks with 2, 4, or 6 hidden layers. To balance model capacity and overfitting, we experimented with various hidden layer sizes and depths to ensure sufficient expressiveness without excessive parameterization.
\ifresubmissionblue}\fi
\ifmodification{\color{red} \st{We tested out different sizes and depth of hidden layers. }}\fi 
Table~\ref{tab:NN_hidden_layers} shows the mean and max testing error of different hidden layers sizes and depths. We choose 3 hidden layers with 200 nodes each which yields the best results.

\begin{table}[htbp]
\centering
\begin{tabular}{c c c c c }
\hline
 & [100,100] & [200,200] & [600,600] & [200,200,200] \\
\hline
Mean & 0.34 & 0.27 & 0.27 & 0.25 \\
Max & 0.91 & 0.71 & 0.67 & 0.65\\
\hline
\hline
\end{tabular}
% \caption{Tool frame position error statistics comparison on the training data, using either numerical (Num) or analytical (ana) gradient matrix. }
\caption{Mean and max testing position error (mm) of different hidden layer structures.}
\label{tab:NN_hidden_layers}
\end{table}

\begin{figure}[ht]
    \centering
    \includegraphics[width=0.30\textwidth]{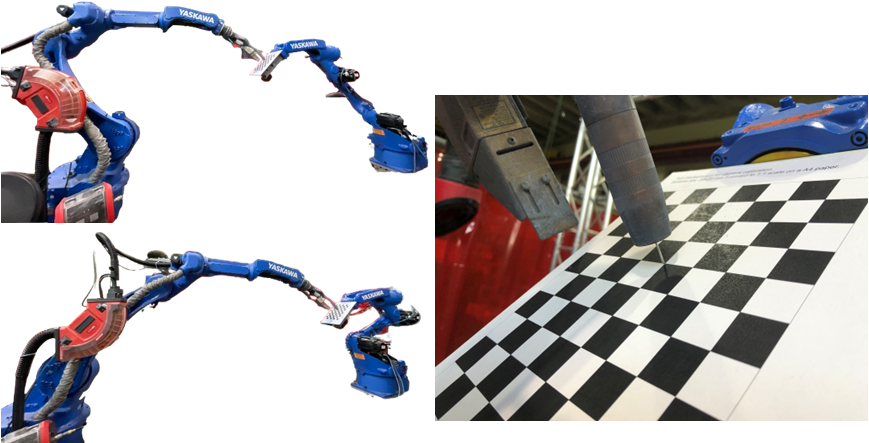}
    \caption{\ifresubmissionblue{\color{blue}\fi  To evaluate performance without using the motion capture system, two robots are jogged to five different configurations, ranging from inward to outstretched, while maintaining contact at the same physical point. \ifresubmissionblue}\fi }
    \label{fig:dual_valid}
\end{figure}

\ifresubmissionblue{\color{blue}\fi 
\noindent{{\em d)} \bf Dual-Robot Validation} 
% For performance comparison without using the motion capture system, we command both robots to have a fixed relative tool position while changing the robot poses from inward to outstretched.
For comparison without motion capture, we command both robots maintaining a fixed relative tool position while their poses varied from inward to outstretched.
%As an independent verification, we check  the applicability of the configuration-dependent POE parameters in a physical setting without using the motion capture system.  
%we conducted a test in which both robots were jogged to five distinct configurations, ranging from inward to outstretched, while R1 maintained contact with the same relative physical point on R2. 
The setup, shown in Fig.~\ref{fig:dual_valid}, simulates a dual-robot collaborative manufacturing task, such as R1 applying a cold spray tool to a workpiece held by R2. By recording the joint angles of both robots and applying different kinematic parameter sets, we compute the resulting position deviations from the known physical setup. Table~\ref{tab:touch_test} summarizes the mean, standard deviation, and maximum position errors for the nominal, CPA, FBF and AE models for five configurations. The result is consistent with performance evaluation using the motion capture system and confirms that the proposed CDC framework is effective for maintaining high relative positioning accuracy in a real dual-robot setup.
\ifresubmissionblue}\fi

\begin{table}[htbp]
\centering
\begin{tabular}{c c c c}
\hline
 & Mean (mm) & Std (mm) & Max (mm) \\
\hline
Nominal & 1.40 & 0.52 & 2.08 \\
CPA & 1.45 & 0.62 & 2.23 \\
FBF N=7 & 0.52 & 0.21 & 0.81\\
AE N=6 & 0.61 & 0.16 & 0.88\\
\hline
\hline
\end{tabular}
% \caption{Tool frame position error statistics comparison on the training data, using either numerical (Num) or analytical (ana) gradient matrix. }
\caption{\ifresubmissionblue{\color{blue}\fi 
Mean, standard deviation and maximum position error of the dual-robot validation. The result is consistent with the performance evaluation using the motion capture system. \vspace{-2mm}
%s show that the proposed CDC framework is repeatable and effective in preserving relative positioning accuracy in real dual-robot operations. 
\ifresubmissionblue}\fi}
\label{tab:touch_test}
\end{table}
}

\section{Conclusion and Future Work} 
\label{sec:conclusion}

The paper introduces a configuration-dependent kinematic model and calibration framework designed to compensate for the inherent variability in serial robot kinematics due to non-geometric factors as a function of the robot pose. The robot kinematics are modeled using a minimal version of the  POE parameters. Calibration is formulated as a non-linear least squares optimization, aimed at minimizing position and orientation errors across different configuration data clusters using a Jacobian iterative approach. We express the robot kinematics as a function of joint 2 and 3 angles and approximate the function using optimized kinematic parameters at sampled configurations. The results demonstrate that the accuracy of the configuration-dependent kinematic model surpasses that of a single universal kinematic model calibrated at one configuration. 
%Furthermore, the paper illustrates that the calibrated kinematic model enhances accuracy in precision-demanding manufacturing processes.
%There are several avenues for future research. 
We are currently working on the calibration of two robot manipulators concurrently.
%in the set up of dual arm cooperation. 

%%%%%%%%%%%%%%%%%%%%%%%%%%%%%%%%%%%%%%%%%%%%%%%%%%%%%%%%%%%%%%%%%%%%%%
\ifNotAnonymous{
\section*{Acknowledgment}

The authors would like to thank Santiago Paternain, Pinghai Yang, Jeffrey Schoonover, John Wason for their helpful discussion of the project and Glenn Saunders, Terry Zhang, Jinhan Ren for their help on the hardware.

\section*{Funding Data}

% \begin{itemize}
Research was sponsored by the ARM (Advanced Robotics for
Manufacturing) Institute through a grant from the Office of the
Secretary of Defense and was accomplished under Agreement
Number W911NF-17-3-0004. The views and conclusions contained in this document are those of the authors and should
not be interpreted as representing the official policies, either
expressed or implied, of the Office of the Secretary of Defense
or the U.S. Government. The U.S. Government is authorized
to reproduce and distribute reprints for Government purposes
notwithstanding any copyright notation herein.
% \end{itemize}

}\fi
%%%%%%%%%%%%%  BIBLIOGRAPHY  %%%%%%%%%%%%%%%%%%%%%%%%%%%%%%%%%%%%%%%%%

% \nocite{*} %% <=== Delete this line - unless you wish to typeset the entire contents of your .bib file.

\iffull{

\appendices
\section{Gradient Computation}
\label{app:gradient}
First consider the rotation matrix,
\begin{equation}
    R_{0T} = R_{01}\ldots R_{n-1,n} R_{nT}
\end{equation}
where $R_{i-1,i}=e^{h_i^\times q_i}$ and $R_{nT}$ is a constant matrix. From the Euler-Rodrigues Formula, we can write $R_{i-1,i}$ as 
\begin{equation}
    R_{i-1,i}= I + \sin(q_i)(h_i)^\times + (1-\cos(q_i))h_i^\times h_i^\times
\end{equation} 
It follows that 
\begin{subequations}
\setlength{\jot}{2pt}
    \begin{align}
        &\frac{\partial R_{i-1,i}}{\partial \theta_i} = \sin(q_i) \left(\frac{dh_i}{d \theta_i}\right)^\times+ \nonumber \\
        &\quad(1-\cos(q_i))\left(\!\!\!\left(\frac{dh_i}{d \theta_i}\right)^\times (h_i)^\times + (h_i)^\times \left(\frac{dh_i}{d \theta_i}\right)^\times\!\!\right)\\
&      \frac{\partial R_{i-1,i}}{\partial \phi_i} = \sin(q_i) \left(\frac{dh_i}{d \phi_i}\right)^\times+ \nonumber \\
        &\quad (1-\cos(q_i))\left(\!\!\!\left(\frac{dh_i}{d \phi_i}\right)^\times (h_i)^\times + (h_i)^\times \left(\frac{dh_i}{d \phi_i}\right)^\times\!\!\right)
        \end{align}
\end{subequations}
From \eqref{eq:joint_axis_error}, we have 
\begin{subequations}
\setlength{\jot}{2pt}
\begin{align}
    \frac{d h_i}{d \theta_i} &= k_{1,i}^{\times} \rot(k_{1,i},\theta_i) \rot(k_{2,i},\phi_i) \bar h_i\\
        \frac{dh_i}{d \phi_i} &= \rot(k_{1,i},\theta_i) k_{2,i}^{\times}\rot(k_{2,i},\phi_i) \bar h_i.
\end{align}
\end{subequations}
Hence, we can compute the gradient of $R_{0T}$ by $(\theta_i,\phi_i)$:
$$%\begin{equation}
\!\!\! \!\!\!   \frac{\partial R_{0T}}{\partial \theta_{i}} = R_{0,i-1}\frac{\partial R_{i-1,i}}{\partial \theta_i}R_{i,T},\,\,
    \frac{\partial R_{0T}}{\partial \phi_{i}} = R_{0,i-1}\frac{\partial R_{i-1,i}}{\partial \phi_i}R_{i,T}.
$$%\end{equation}
%If $j<i$, then the gradient of $R_{0j}$ w.r.t. $(\theta_i,\phi_i)$ is zero.
% Hence, for $j\ge i$ ($j=n+1$ is the tool frame), we can compute the gradient with respect to the rotational POE parameters as
% \begin{equation}
% \!\!\! \!\!\!   \frac{\partial R_{0j}}{\partial \theta_{i}} = R_{0,i-1}\frac{\partial R_{i-1,i}}{\partial \theta_i}R_{i,j},\,\,
%     \frac{\partial R_{0j}}{\partial \phi_{i}} = R_{0,i-1}\frac{\partial R_{i-1,i}}{\partial \phi_i}R_{i,j}.
% \end{equation}
% \end{subequations}
% If $j<i$, then the gradient of $R_{0j}$ by $(\theta_i,\phi_i)$ is zero.
The position kinematics is given by
\begin{equation}
     p_{0T} = p_{01} + R_{01}p_{12} + ... + R_{0n}p_{nT}.
\end{equation}
From \eqref{eq:link_O}, we have 
\begin{subequations}
\setlength{\jot}{2pt}
\begin{align}
        \frac{dp_{i-1,i}}{d v_i} &= k_{1,i}, \quad
        \frac{dp_{i-1,i}}{d w_i} = k_{2,i}   \\
        \frac{dp_{i,i+1}}{d v_i} &= -k_{1,i}, \quad
        \frac{dp_{i,i+1}}{d w_i} = -k_{2,i}.   
\end{align}
\end{subequations}
It follows 
$$%\begin{equation}
\setlength{\jot}{2pt}
    \begin{aligned}
\!\!\!        \frac{dp_{0T}}{dv_i}&= (R_{0,i-1}-R_{0,i})k_{1,i},\,
        \frac{dp_{0T}}{dw_i}= (R_{0,i-1}-R_{0,i})k_{2,i}\\
\!\!\! \frac{dp_{0T}}{d\theta_i} &= R_{0,i-1}\frac{dR_{i-1,i}}{d\theta_i} p_{iT},\, 
        \frac{dp_{0T}}{d\phi_i} = R_{0,i-1}\frac{dR_{i-1,i}}{d\phi_i} p_{iT}.
    \end{aligned}
$$%\end{equation}
where $p_{iT}=p_{i,i+1}+\ldots+R_{in}p_{nT}$.
Using the definition of the parameter vectors in \eqref{eq:paramvec}, we have %the position gradient as 
\begin{equation}
\setlength{\jot}{2pt}
\begin{aligned} 
    \frac{\partial p_{0T}}{\partial\ERP} &= \begin{bmatrix}
        \frac{dp_{0T}}{dv_1} &         \frac{dp_{0T}}{dw_1} & \ldots &         \frac{dp_{0T}}{dv_n} &         \frac{dp_{0T}}{dw_n} 
    \end{bmatrix} \\
        \frac{\partial p_{0T}}{\partial\ERH} &= \begin{bmatrix}
        \frac{dp_{0T}}{d\theta_1} &   \frac{dp_{0T}}{d\phi_1} & \ldots &         \frac{dp_{0T}}{d\theta_n} &         \frac{dp_{0T}}{d\phi_n} 
    \end{bmatrix}. 
\end{aligned}
\end{equation}
For the orientation, let $\psi$ be any 3-parameter representation of $SO(3)$ and $J_\psi$ be the representation Jacobian, i.e., $\dot \psi = J_\psi \omega$.  Then the gradient of $\psi$ with respect to the parameter vectors is given by
\begin{equation}
\setlength{\jot}{2pt}
\begin{aligned}
&\frac{\partial \psi}{\partial \ERH}
= J_\psi\bigl[
        (\frac{dR_{0T}}{d\theta_1}R_{0T}\tr)^\vee, %\,\,  (\frac{dR_{0T}}{d\phi_1}R_{0T}\tr)^\vee \,\, 
        \ldots ,
        %\\
%&\qquad  \qquad  \qquad     
%(\frac{dR_{0T}}{d\theta_n}R_{0T}\tr)^\vee, \,\,   
(\frac{dR_{0T}}{d\phi_n}R_{0T}\tr)^\vee\bigr]
\end{aligned}   
\end{equation}
and $\frac{\partial \psi}{\partial\ERP} = 0$, 
where $(\cdot)^\vee$ converts a skew symmetric matrix to a vector (inverse of $(\cdot)^\times$).

}\fi

\bibliographystyle{IEEEtran}
\bibliography{bib}

\end{document}